\PassOptionsToPackage{table,xcdraw,dvipsnames}{xcolor}
\documentclass[lettersize,journal]{IEEEtran}
\usepackage{amsmath,amsfonts}
\usepackage{algorithmic}
\usepackage{algorithm}
\usepackage{array}
\usepackage[caption=false,font=normalsize,labelfont=sf,textfont=sf]{subfig}
\usepackage{textcomp}
\usepackage{stfloats}
\usepackage{url}
\usepackage{verbatim}
\usepackage{graphicx}
\usepackage{cite}
\usepackage{tikz}
\usepackage{forest}
\usepackage{amssymb}
\usepackage{pifont}
\usetikzlibrary{trees,positioning,shapes,shadows,arrows.meta}
\usepackage{makecell}
\usepackage{txfonts}
\usepackage{multirow}
\usepackage{subcaption}
\usepackage{makecell}
\usepackage{tabularx}
\usepackage[capitalise]{cleveref}
\usepackage{subfig}

\newcommand{\cmark}{\ding{51}} % Check mark
\newcommand{\xmark}{\ding{55}} % Cross mark
\usepackage{booktabs}

\newcommand{\revise}[1]{\textcolor{black}{#1}}
\newcommand{\rev}[1]{\textcolor{black}{#1}}
\begin{document}

\title{A Comprehensive Survey of Knowledge-Based Visual Question Answering Systems: The Lifecycle of Knowledge in Visual Reasoning Task}

\author{Jiaqi Deng, Zonghan Wu, Huan Huo, Guandong Xu~\IEEEmembership{Senior Member, IEEE}

        % <-this % stops a space
% <-this % stops a space
\thanks{Manuscript received April 19, 2021; revised August 16, 2021.}
\thanks{Jiaqi Deng and Huan Huo are with the University Technology of Sydney, Australia (email: jiaqi.deng@student.uts.edu.au, huan.huo@uts.edu.au)}
\thanks{Zonghan Wu is with the East China Normal University, China (email: zhwu@sem.ecnu.edu.cn)}
\thanks{Guandong Xu is with the Education University of Hong Kong, China (email: gdxu@eduhk.hk)}
}

\markboth{Journal of \LaTeX\ Class Files,~Vol.~14, No.~8, August~2021}%
{Shell \MakeLowercase{\textit{et al.}}: Survey of Knowledge-based Visual Question Answering Systems: Knowledge Integration Across System Stages}


\maketitle

\begin{abstract}
Knowledge-based Visual Question Answering (KB-VQA) extends general Visual Question Answering by requiring external knowledge beyond the provided visual and textual inputs, facilitating more complex real-world applications. KB-VQA introduces unique challenges, including the alignment of heterogeneous information from diverse modalities and sources, the retrieval of relevant knowledge from large-scale and noisy repositories, and the execution of complex reasoning to infer answers from the combined context. With the advancement of large language models, KB-VQA systems have undergone a notable transformation, where LLMs serve as powerful knowledge repositories, retrieval-augmented generators and strong reasoners. Despite substantial progress, there is a lack of a recent, systematic survey that organizes and reviews the evolving landscape of existing KB-VQA methods. This survey aims to fill this gap by establishing a structured taxonomy of KB-VQA approaches and decomposing mainstream systems into three fundamental stages: knowledge representation, knowledge retrieval, and knowledge reasoning. Through an examination of existing techniques employed at each stage, this survey identifies persistent challenges and outlines promising future research directions, providing a foundation for advancing KB-VQA models and their applications.
\end{abstract}


\section{Introduction}
\begin{figure}[t]
    \centering
   \centerline{\includegraphics{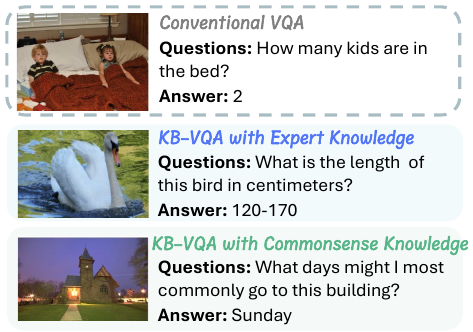}}
    \caption{A comparison between conventional VQA question and KB-VQA questions with different knowledge requirements.}
    \label{examples}
\end{figure}

\bstctlcite{IEEEexample:BSTcontrol}
\IEEEPARstart{V}{isual} question answering (VQA)~\cite{Agrawal2017VQA:Answering} has long stood as one of the most compelling challenges residing at the intersection of computer vision and natural language processing. While conventional VQA systems aim to answer questions directly from visual context, Knowledge-based VQA (KB-VQA)~\cite{Marino2019OK-VQA:Knowledge} presents a particular extension, where it necessitates not only understanding the visual and textual inputs but also effective incorporation and reasoning over knowledge repositories. As demonstrated in \revise{\cref{examples}}, conventional VQA involves questions that can be directly answered by processing and describing an image, exemplified by questions like ``\texttt{How many kids are in the bed?}''. In contrast, KB-VQA often requires domain-specific understanding. For instance, answering the question ``\texttt{What days might I most commonly go to this building?}'' involves commonsense knowledge that ``\texttt{people often go to church for praying on Sundays}''. Similarly, answering ``\texttt{What is the length of this bird?}'' is not possible without expert knowledge about bird biology. Through incorporating commonsense knowledge, open-world knowledge and domain-specific knowledge, KB-VQA systems are crucial for applications in various domains, such as medical diagnosis~\cite{Xia2025MMed-RAG:Models, Naeem2024Path-RAG:Answering} and interactive education~\cite{Kembhavi2017AreComprehension}.

The integration of knowledge in KB-VQA brings unique challenges that are absent in conventional VQA. First, multimodal \textbf{knowledge representation} remains a fundamental challenge. Unlike traditional VQA, where the model operates solely on image and text pairs, KB-VQA systems must align retrieved knowledge with both visual and textual modalities at a fine-grained level. This requires effective cross-modal representation learning mechanisms to align knowledge representations and high-dimensional image-question features. \revise{Second, as systems need to navigate large and heterogeneous knowledge sources, effective \textbf{knowledge retrieval} is essential, where identifying pertinent facts while filtering out noise can be particularly challenging for ambiguous or context-dependent queries. Finally, \textbf{knowledge reasoning} represents another fundamental challenge in synthesizing accurate responses. The mere retrieval of relevant information does not ensure successful utilization; rather, KB-VQA demands sophisticated reasoning to integrate disparate information and uncover latent dependencies not explicitly present in the input. In complex scenarios, models might also be required to effectively process very long knowledge contexts and perform multi-step inference.}

Existing survey papers on general VQA~\cite{Zakari2025VQADirection, Srivastava2021VisualAnalysis, Kuang2025NaturalSurvey} and multimodal learning techniques~\cite{Jabeen2023ALearning, Wu2023MultimodalSurvey} focus primarily on the general integration of multimodal data. As discussed above, KB-VQA possesses distinct challenges that cannot be appropriately addressed by frameworks developed for general VQA tasks. \revise{The rise of Large Language Models (LLMs) has further reshaped the field by acting as knowledge sources with its large-scale implicit knowledge or performing knowledge-focus tasks with well-designed prompt-based mechanisms, offering new solutions to the long-standing KB-VQA challenges. This survey focuses on KB-VQA methods that explicitly leverage LLMs as knowledge access and integration, rather than on direct application of general-purpose LLMs to VQA.}
\begin{figure*}[tb]
    \centering  \includegraphics[width=\textwidth]{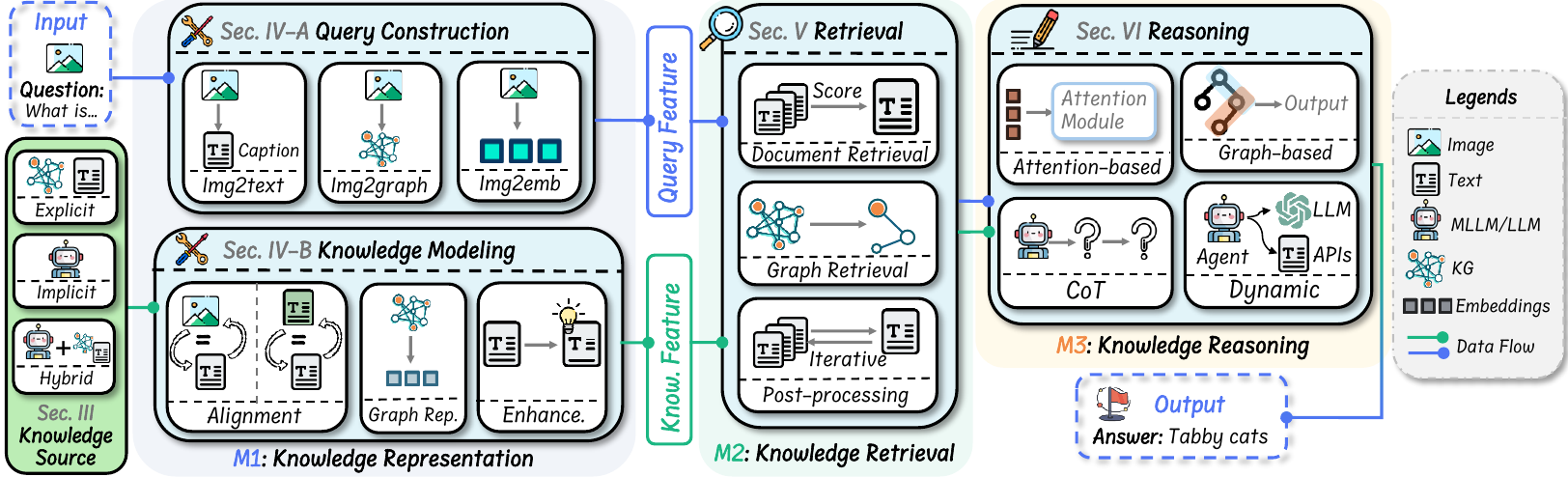}
    \caption{An overview of the generalized pipeline for KB-VQA, where each section provides an abstract representation of the main classifications. In this figure, ``Knowledge" is abbreviated as ``Know." and ``Chain-of-Thought" is abbreviated as ``CoT".}
    \label{guideline}
\end{figure*}

\revise{Additionally, we acknowledge the growing importance of KB-VQA in specialized fields such as medical imaging and geospatial analysis, where domain-specific adaptations are often required. For example, medical VQA systems often feature specialized image encoders that work with different medical imaging modalities~\cite{Lin2023MedicalSurvey}. These areas have been covered in dedicated literature~\cite{Lin2023MedicalSurvey,Wang2024EarthVQA:Answering,Zi2025RSVLM-QA:Answering}. With a central focus on general-purpose KB-VQA systems, our paper serves as a complementary guide to the core knowledge-centric architectures that those specialized systems build upon. }

As the field rapidly advances, a comprehensive overview of existing KB-VQA studies is urgently needed - one that focuses specifically on knowledge representation, knowledge retrieval, and knowledge reasoning techniques, while also capturing emerging trends in the LLM-driven progression of KB-VQA methods. To the best of our knowledge, this is the first comprehensive survey to provide a holistic overview of this evolving field. It aims to systematically review notable KB-VQA approaches, with the following key contributions:

\begin{itemize}
    \item We propose a structured taxonomy framework (\revise{\cref{fig:lit_surv}}) for categorizing existing KB-VQA methods to highlight their distinctions and commonalities.
    \item We analyze the architectural designs and knowledge incorporation mechanisms at each stage of KB-VQA systems to fully understand the knowledge lifecycle.
    \item We identify key challenges and limitations in current approaches and offer an outlook for future advancements.
\end{itemize}

By conducting a structured analysis, we seek to provide valuable insights into the development of KB-VQA and highlight promising directions for future research. The works covered in this survey are collected from top-tier conferences and journals, ensuring a rigorous and representative overview of the field. This paper is organized as follows. \cref{def} defines the KB-VQA task and its setting. \cref{know_sources} examines various knowledge sources leveraged by KB-VQA systems and their association. \cref{representation} - \ref{knowledge retreival section} provide a comprehensive overview of the main components of KB-VQA, including knowledge representation, retrieval, and reasoning. \cref{benchmarks} summarizes popular benchmarks and discusses performance comparison of representative methods, while \cref{challenges} outlines potential future research directions.

% KB-VQA unique challenges needed to be highlighted in the graph.

\section{Problem Definitions}\label{def}
\noindent KB-VQA involves answering natural language questions about images that require external knowledge. Unlike general VQA, which relies solely on visual content and textual queries, KB-VQA incorporates additional knowledge sources to generate more informed answers. We generalize the standard KB-VQA framework into three distinct phases, each with a clear definition: \revise{Knowledge Representation}, Knowledge Retrieval, and Knowledge Reasoning, as shown in \revise{\cref{guideline}}.

\textbf{Knowledge Representation ($\mathcal{M}_1$)}: the initial phase processes the input image-question pair to generate feature representations that facilitate knowledge retrieval and answer generation. This phase also involves representing external knowledge sources in a compatible format and aligning them with the visual and textual embeddings. 

\textbf{Knowledge Retrieval ($\mathcal{M}_2$)}: \revise{This phase identifies and retrieves relevant knowledge based on the context representation obtained from $\mathcal{M}_1$.} It involves querying from explicit knowledge bases (KBs) (e.g., Knowledge Graphs (KGs) or textual documents) or implicit knowledge from large pre-trained models by designed prompts.

% definition of reasoning to be clear.
\textbf{Knowledge Reasoning ($\mathcal{M}_3$)}: \revise{This final phase integrates the multimodal representation from $\mathcal{M}_1$ with the retrieved knowledge from $\mathcal{M}_2$ to infer the final answer.} It applies reasoning mechanisms such as attention-based fusion, graph-based inference to utilize the knowledge and context effectively.

\section{Knowledge sources} \label{know_sources}

\begin{table*}[t]
    \centering
    \caption{\revise{Comparison of knowledge sources in KB-VQA.}}
    \resizebox{\textwidth}{!}{
    \begin{tabular}{l|>{\raggedright\arraybackslash}p{3.5cm}|>{\raggedright\arraybackslash}p{3.5cm}|>{\raggedright\arraybackslash}p{3.5cm}}
        \toprule
        \textbf{Knowledge Source} & \textbf{Explicit Knowledge} & \textbf{Implicit Knowledge} & \textbf{Hybrid Knowledge} \\
        \midrule
        \textbf{Description} & Structured KGs or unstructured corpus & Pre-trained LLMs or MLLMs
        & Combination of explicit knowledge and LLMs \\
         % & \parbox{3.5cm}{\raggedright\begin{itemize}\item Requires external retrieval\end{itemize}} 
         % & \parbox{3.5cm}{\raggedright\begin{itemize}\item Prompt-based implicit retrieval\end{itemize}} 
         % & \parbox{3.5cm}{\raggedright\begin{itemize}\item Adaptive retrieval with LLM reasoning\end{itemize}} \\
        \midrule
        \textbf{Ease of Update} & \cellcolor{YellowGreen!25}\cmark (High, knowledge update is straightforward) & \cellcolor{Maroon!25}\xmark (Limited, knowledge is internalized) & \cellcolor{YellowGreen!25}\cmark (The explicit KB can be easily updated) \\
        \midrule
        \textbf{Access to Open-world Knowledge} & \cellcolor{Maroon!25}\xmark (Limited, relies on similarity score or keyword matching) & \cellcolor{YellowGreen!25}\cmark (High, prompting provides flexible knowledge interface) & \cellcolor{YellowGreen!25}\cmark (High, provided by implicit knowledge) \\
        % \midrule
        % \textbf{Zero-Shot Capability} & \cellcolor{Maroon!25}\xmark (Requires retrieval and fine-tuning) & \cellcolor{YellowGreen!25}\cmark (Strong zero-shot reasoning) & \cellcolor{YellowGreen!25}\cmark (Can retrieve and reason effectively) \\
        \midrule
        \textbf{Hallucination Risk} & \cellcolor{YellowGreen!25}\cmark (Low, knowledge is grounded) & \cellcolor{Maroon!25}\xmark (High, may generate incorrect facts) & \cellcolor{yellow!20}$\triangle$ (Moderate, depends on retrieval accuracy) \\    
        \midrule
        \textbf{Interpretability} & \cellcolor{YellowGreen!25}\cmark (High, human readable) & \cellcolor{Maroon!25}\xmark (Low, embedded in model parameters) & \cellcolor{YellowGreen!25}\cmark (Combines structured facts and reasoning) \\
        \midrule
        \textbf{Representative Methods} & TRiG, RA-VQA, KG-Aug & PICa, PnP-VQA, GeReA & KAT, REVIVE, FLMR, AVIS \\
        \bottomrule

    \end{tabular}}
    \label{knowledge_comparison}
\end{table*}
\revise{This section systematically compares different forms of knowledge used in KB-VQA systems: explicit, implicit, and hybrid type. We organize representative approaches according to how knowledge is represented, accessed, and integrated into KB-VQA systems.
\cref{knowledge_comparison} presents a summary and comparison of key characteristics of knowledge sources, which highlights that various knowledge sources fundamentally shape how KB-VQA systems interact with the knowledge.}
\subsection{Explicit Knowledge}
\revise{Explicit knowledge in KB-VQA refers to information sourced from external, human-curated repositories that are accessed through \textcolor{black}{\textsc{Retrieval-based Systems}}. We broadly categorize explicit knowledge into structured and unstructured knowledge; this distinction dictates how KB-VQA frameworks interact with the external knowledge.}

% \revise{Structured knowledge is typically organized as KGs (e.g. ConceptNet~\cite{Speer2016ConceptNetKnowledge} and WikiData~\cite{Vrandecic2014Wikidata}), where facts are represented as triples \texttt{(head entity, relation, tail entity)}.} \revise{Early KB-VQA systems transform image and question inputs into structured queries to retrieve relevant facts from KGs~\cite{Wu2018ImageKnowledge}. While this paradigm offers interpretability, it is constrained by predefined schemas and limited adaptability to novel questions. To mitigate this limitation, KG-Aug~\cite{Li2020BoostingAggregation} introduces a context-aware knowledge aggregation framework that relaxes strict query matching and enhances visual feature representations with external KGs.}

\revise{Structured knowledge in KB-VQA is typically organized as KGs (e.g. ConceptNet~\cite{Speer2017ConceptNetKnowledge} and WikiData~\cite{Vrandecic2014Wikidata}), where facts are formalized as relational triples \texttt{(head entity, relation, tail entity)}. This representation enables precise symbolic reasoning and high interpretability, as reasoning paths can be explicitly traced through graph traversal~\cite{Wu2018ImageKnowledge}. However, systems that rely exclusively on explicit knowledge are often constrained by schema rigidity and data sparsity. Operating under a “closed-world” assumption, they struggle to accommodate long-tail concepts beyond predefined schemas. Moreover, accurately grounding noisy visual regions or natural-language queries to specific knowledge graph nodes remains a non-trivial challenge and continues to attract active research interest~\cite{Li2020BoostingAggregation}.}

\revise{Unstructured knowledge leverages extensive corpora of free-form information, ranging from Wikipedia articles~\cite{Lin2023Fine-grainedAnswering} to web-scale search results~\cite{Luo2021Weakly-SupervisedAnswering}. Compared to structured KGs, unstructured sources utilize the inherent flexibility of natural language to provide superior knowledge coverage. Additionally, compared to symbolic matching, they offers more robust semantic alignment through multimodal features~\cite{Gao2022Transform-Retrieve-Generate:Answering}. Nevertheless, large-scale corpora introduces significant information noise and retrieval overhead. Due to the presence of redundant or contradictory knowledge, effective KB-VQA systems must employ robust and flexible retrieval mechanisms. Therefore, active research on integration of unstructured knowledge to KB-VQA are generally centered around highly optimized similarity search algorithms~\cite{Lin2023Fine-grainedAnswering}, sophisticated training strategies~\cite{Deng2025EnablingAnswering, Hong2025Knowledge-basedFiltering} and post-processing techniques~\cite{Hu2023AVIS:Agent,Zhou2023PromptingVQA}.}

\subsection{Implicit Knowledge}
\revise{The scope of knowledge sources has recently expanded with the advancement of large pre-trained models. Through extensive pretraining on massive datasets, LLMs and Multimodal Large Language Models (MLLMs) have acquired vast amounts of knowledge that is stored in model parameters, which we refer to as implicit knowledge. This development has given rise to \textcolor{black}{\textsc{prompt-based systems}}, where pre-trained models are explicitly incorporated into the KB-VQA pipeline to access and integrate implicit knowledge via carefully designed prompts, rather than being applied directly as generic VQA solvers.}

For example, PICa~\cite{Yang2022AnVQA} is the first work that leverages frozen GPT-3~\cite{Brown2020LanguageLearners} for KB-VQA task. It feeds a few in-context examples alongside the target question to the LLM, enabling which to learn and apply patterns for answering image-based questions through demonstration. PnP-VQA~\cite{Tiong2022Plug-and-PlayTraining} is another impressive work that proposes a modular framework without training that converts the image into question-relevant captions to activate the capacity of UnifiedQAv2~\cite{Khashabi2020UNIFIEDQA:System}. However, PromptCap~\cite{Hu2023PromptCap:Captioning} \revise{argues that converting the image into generic text can exclude essential visual information.} Therefore, they introduce a question-guided image caption model to serve as a bridge between images and language models.

MLLMs such as InstructBLIP~\cite{Dai2023InstructBLIP:Tuning}, BLIP2~\cite{Li2023BLIP-2:Models}, and PaLM-E~\cite{Driess2023PaLM-E:Model} have shown strong capabilities in aligning visual and textual modalities, which suggests that multimodal pretraining enables more direct access to implicit multimodal knowledge. Building on this capability, GeReA~\cite{Ma2024GeReA:Answering} leverages MLLMs as implicit knowledge sources by prompting them with selective in-context examples and further improves the answering accuracy. \revise{Compared to explicit knowledge sources, implicit knowledge enables easier access to open-world knowledge and reduces reliance on external retrieval operations. However, because this type of knowledge is internalized as model parameters, the acquisition can be less transparent and might suffer from hallucinations~\cite{Mallen2023WhenMemories,Min2023FActScore:Generation}. These limitations highlight fundamental trade-offs between flexibility and accuracy of retrieved knowledge, which motivates ongoing research on hybrid approaches that aim to strike a balance.}

\subsection{Hybrid Knowledge}
\revise{To leverage the complementary strengths of both explicit and implicit knowledge, recent KB-VQA research has explored hybrid knowledge sources, which combines both implicit and explicit knowledge sources. With access to both types of knowledge, hybrid knowledge benefits from the interpretability and low hallucination risk of explicit knowledge, as well as the access to open-world knowledge provided by LLMs. For hybrid knowledge, studies have been centered around the effective cross interaction between implicit and explicit knowledge.} Earlier works such as KAT~\cite{Gui2022KAT:Vision-and-Language} uses a contrastive-learning-based module to retrieve knowledge entries from an explicit KB, and uses GPT-3~\cite{Brown2020LanguageLearners} to retrieve implicit knowledge with supporting evidence. Then the knowledge will be integrated jointly by a reasoning module and the decoder. REVIVE~\cite{Lin2022REVIVE:Answering} further highlights the importance of regional information when retrieving both implicit and explicit types of knowledge. 

% RAG paradigm % The difference between classic hybrid methods and RAG-based methods.
\revise{More recent work extends this integration under the \textcolor{black}{\textsc{RAG methods}}, where LLMs are not just standalone knowledge containers but also active reasoner that coordinate retrieval and generation.}
RA-VQAv2~\cite{Lin2022RetrievalKnowledge} exploits LLMs to comprehend question context and perform the retrieval of external knowledge to augment the generation process. AVIS~\cite{Hu2023AVIS:Agent} proposes an LLM-powered planner to dynamically determine either direct answering or search over external knowledge for relevant answers. Similarly,~\cite{Wang2023TowardsAnswering} introduces a multi-agent framework that enhances VLMs through collaborative reasoning between three specialized agents: a VLM-based Responder, an LLM-powered Seeker, and an Integrator, collectively enabling dynamic combination of implicit and explicit knowledge. \rev{Integrating LLMs with external knowledge bases enables more adaptive KB-VQA systems and reflects a broader shift toward agentic, multimodal RAG architectures. This trend highlights how multimodal RAG provides new architectural solutions for KB-VQA. Specifically, it moves beyond reliance on textual knowledge to dynamically leverage heterogeneous sources, including visual and auditory modalities. In this setting, access to more diverse, open-world knowledge improves the scalability and adaptability of KB-VQA systems.}

\begin{figure*}[htbp]
    \centering
    \tikzset{
    basic/.style  = {draw, text width=3cm, align=center, font=\sffamily, rectangle},
    root/.style   = {basic, rounded corners=2pt, thin, align=center, fill=green!30},
    onode/.style = {basic, thin, rounded corners=2pt, align=center, fill=green!60,text width=3cm,},
    tnode1/.style = {basic, thin, align=left, fill=cyan!10!, text width={\textwidth - 29mm - 16mm - 4mm - 2mm-6*0.333em-1mm}, align=left, font=\footnotesize},
    tnode2/.style = {basic, thin, align=left, fill=yellow!10!, text width={\textwidth - 23mm - 2mm-4*0.333em-1mm}, align=left, font=\footnotesize},
    tnode3/.style = {basic, thin, align=left, fill=green!10!, text width={\textwidth - 23mm - 2mm-4*0.333em-1mm}, align=left, font=\footnotesize},
    tnode4/.style = {basic, thin, align=left, fill=green!10!, text width={\textwidth - 23mm - 27mm - 4mm - 2mm-6*0.333em-1mm}, align=left, font=\footnotesize},
    xnode1/.style = {basic, thin, rounded corners=2pt, align=center, fill=cyan!30!, text width=2.9cm, font=\footnotesize},
    xnode11/.style = {basic, thin, rounded corners=2pt, align=center, fill=cyan!30!, text width=1.6cm, font=\footnotesize},
    xnode2/.style = {basic, thin, rounded corners=2pt, align=center, fill=yellow!30!, text width=2.3cm, font=\footnotesize},
    xnode3/.style = {basic, thin, rounded corners=2pt, align=center, fill=lime!30!, text width=2.3cm, font=\footnotesize},
    xnode31/.style = {basic, thin, rounded corners=2pt, align=center, fill=lime!30!, text width=2.7cm, font=\footnotesize},
    wnode/.style = {basic, thin, align=left, fill=pink!10!red!10, text width=6.5em}

}
    \begin{minipage}{\textwidth}
    \begin{minipage}{\textwidth}
    \begin{forest} for tree={
    grow=east,
    growth parent anchor=west,
    s sep=1mm,
    parent anchor=east,
    child anchor=west,
    anchor=parent,
    edge path={\noexpand\path[\forestoption{edge},draw] 
         (!u.parent anchor) -- +(6pt,0pt) |-  (.child anchor) 
         \forestoption{edge label};}
}
        [Query construction, xnode11,  l sep=4mm,
        [Image-to-graph, xnode1, l sep=2mm,[Mucko\protect~\cite{Zhu2020Mucko:Answering};\,GRUC\protect~\cite{Yu2020Cross-modalAnswering};\,VQA-GNN\protect~\cite{Wang2023VQA-GNN:Answering};\,
        \protect~\cite{Heo2022HypergraphAnswering};\,, tnode1, l sep=16mm]],
        [Image-to-embed, xnode1, l sep=2mm,
        [CFM\protect~\cite{Nguyen2022Coarse-to-FineAnswering};\,FLMR\protect~\cite{Lin2023Fine-grainedAnswering}; \textcolor{brown}{GeReA}\protect~\cite{Ma2024GeReA:Answering}
        , tnode1, l sep=16mm]],
        [Image-to-text, xnode1, l sep=2mm,[\protect~\cite{Wu2016AskSources};\,\protect~\cite{Wu2018ImageKnowledge};\,QANet\protect~\cite{Li2019VisualComprehension};\,KAT\protect~\cite{Gui2022KAT:Vision-and-Language};\, TRiG\protect~\cite{Gao2022Transform-Retrieve-Generate:Answering};\,MAVEx\protect~\cite{Wu2022Multi-ModalVQA};\, REVIVE\protect~\cite{Lin2022REVIVE:Answering};\,\textcolor{brown}{PICa}\protect~\cite{Yang2022AnVQA};\,
        \textcolor{brown}{PnPVQA}\protect~\cite{Tiong2022Plug-and-PlayTraining};\, \textcolor{brown}{LAMOC}\protect~\cite{Du2023Zero-shotFeedback};\,
         \textcolor{brown}{PromptCap}\protect~\cite{Hu2023PromptCap:GPT-3};\, \textcolor{brown}{Img2LLM}\protect~\cite{Guo2023FromModels};\,\textcolor{brown}{\protect~\cite{Wang2023FillingQuestions}};\, CBM\protect~\cite{Salaberria2023ImageAnswering}, tnode1, l sep=16mm]]
         ]
    \end{forest}
    \vspace{-3mm}
    \end{minipage}\\
    \begin{minipage}{\textwidth}

    \begin{forest} for tree={
    grow=east,
    growth parent anchor=west,
    s sep=1mm,
    parent anchor=east,
    child anchor=west,
    anchor=parent,
    edge path={\noexpand\path[\forestoption{edge},draw] 
         (!u.parent anchor) -- +(6pt,0pt) |-  (.child anchor) 
         \forestoption{edge label};}
}
            [Knowledge modeling, xnode11,  l sep=4mm,
            [Knowledge enhancement, xnode1, l sep=2mm,[\protect~\cite{Hao2024KnowledgeVQA};\,\protect~\cite{Adjali2023ExplicitEntities};\,\protect~\cite{Sun2023BreakingVQA};\,\textcolor{brown}{SelTDA}\protect~\cite{Khan2023Q:Images};\,\textcolor{brown}{PLLMKI}\protect~\cite{Hu2024PromptingAnswering}, tnode1, l sep=16mm]],
            [Graph-based knowledge representation, xnode1, l sep=2mm,[VKMN\protect~\cite{Su2018LearningAnswering};\,KRISP\protect~\cite{Marino2021KRISP:VQA};\,KM$^4$\protect~\cite{Zheng2021KM4:Modulation};\,MuKEA\protect~\cite{Ding2022MuKEA:Answering};\,Mucko\protect~\cite{Chen2022LaKo:Injection};\,DMMGR\protect~\cite{Li2022DynamicAnswering};\,\protect~\cite{Heo2022HypergraphAnswering};\,REAVL\protect~\cite{Rao2023Retrieval-basedPre-training};\\KECVQG\protect~\cite{Chen2023DeconfoundedInference};\, HCNMN\protect~\cite{Zhang2023TowardKnowledge};\,HyperMR\protect~\cite{Wang2024HyperMR:Answering};\,CMLR\protect~\cite{Wang2024Cross-modalityAnswering};\,
                OWKRL\protect~\cite{Zheng2024KnowledgeReasoning}, tnode1, l sep=16mm]], 
            [Knowledge alignment, xnode1, l sep=2mm,
                [KVLBERT\protect~\cite{Song2021KVL-BERT:Reasoning};\,LaKo\protect~\cite{Chen2022LaKo:Injection};\,
                VLCBERT\protect~\cite{Ravi2023VLC-BERT:Knowledge};\,FLMR\protect~\cite{Lin2023Fine-grainedAnswering};\,DEDR\protect~\cite{Salemi2023AAnswering};\,MCUD\protect~\cite{Hu2024AVQA};\,CVLM\protect~\cite{Li2024CognitiveAlignment}, tnode1, l sep=16mm]]]
        \end{forest}
    \end{minipage}
    \vspace{-2mm}
    \caption*{\footnotesize(a) Knowledge Representation}
    \vspace{3mm}
    \end{minipage}
    \\
    \begin{minipage}{\textwidth}
    \begin{minipage}{\textwidth}
        \begin{forest} for tree={
    grow=east,
    growth parent anchor=west,
    parent anchor=east,
    child anchor=west,
    anchor=parent,
    edge path={\noexpand\path[\forestoption{edge},draw] 
         (!u.parent anchor) -- +(4pt,0pt) |-  (.child anchor) 
         \forestoption{edge label};}
    }
        [Document retrieval, xnode2, l sep=2mm,
        [\protect~\cite{Luo2021Weakly-SupervisedAnswering};\,MAVEx\protect~\cite{Wu2022Multi-ModalVQA};\,UnifER\protect~\cite{Guo2022AVQA};\,RAVQA\protect~\cite{Lin2022RetrievalKnowledge};\,FLMR\protect~\cite{Lin2023Fine-grainedAnswering};\,DEDR\protect~\cite{Salemi2023AAnswering};\,\protect~\cite{Reichman2023Cross-ModalAnswering};\,\protect~\cite{Lerner2024Cross-ModalAnswering};\,PreFLMR\protect~\cite{Lin2024PreFLMR:Retrievers};\,GeMKR\protect~\cite{Long2024GenerativeModels};\,CEIK\protect~\cite{Wang2024CoordinatingVQA};\,
        UniRVQA\protect~\cite{Deng2025EnablingAnswering};\,VLM-PRF\protect~\cite{Hong2025Knowledge-basedFiltering};\,BayesRAG\protect~\cite{Li2026BayesRAG:Generation}, tnode2, l sep=16mm]]
        \end{forest}
        \vspace{-3mm}
    \end{minipage}\\
    \begin{minipage}{\textwidth}
        \begin{forest} for tree={
    grow=east,
    growth parent anchor=west,
    parent anchor=east,
    child anchor=west,
    anchor=parent,
    edge path={\noexpand\path[\forestoption{edge},draw] 
         (!u.parent anchor) -- +(4pt,0pt) |-  (.child anchor) 
         \forestoption{edge label};}
    }   
        [Graph retrieval, xnode2, l sep=2mm,
        [Straight to the Facts\protect~\cite{Narasimhan2018StraightAnswering};\, KG-Aug\protect~\cite{Li2020BoostingAggregation};\, VQA-GNN\protect~\cite{Wang2023VQA-GNN:Answering};\, RR-VEL\protect~\cite{You2023AAnswering};\,
        DSAMR\protect~\cite{Sun2024DSAMR:Answering},tnode2, l sep=16mm]]
        \end{forest}
        \vspace{-3mm}
    \end{minipage}\\
    \begin{minipage}{\textwidth}
        \begin{forest} for tree={
    grow=east,
    growth parent anchor=west,
    parent anchor=east,
    child anchor=west,
    anchor=parent,
    edge path={\noexpand\path[\forestoption{edge},draw] 
         (!u.parent anchor) -- +(4pt,0pt) |-  (.child anchor) 
         \forestoption{edge label};}
    }
        [Post-processing, xnode2, l sep=2mm,
            [\protect~\cite{Wu2018ImageKnowledge};\, PROOFREAD\protect~\cite{Zhou2023PromptingVQA};\,\protect~\cite{Wang2023FillingQuestions};\, LV$^2$-Net\protect~\cite{Liu2024LookGeneration};\, AVIS\protect~\cite{Hu2023AVIS:Agent};\,SIRI\protect~\cite{Wang2023TowardsAnswering};\,Wiki-LLaVa\protect~\cite{Caffagni2024Wiki-LLaVA:LLMs};\,DIETCOKE\protect~\cite{Li2024DiversifyVQA};\,MMV\protect~\cite{Xu2024Multi-ModalAnswering};\,Boter\protect~\cite{Hao2024Boter:VQA},tnode2, l sep=16mm]]
        \end{forest}
        
    \end{minipage}
    \vspace{-4mm}
    \caption*{\footnotesize(b) Knowledge Retrieval}
    \vspace{3mm}
    \end{minipage}\\
        \begin{minipage}{\textwidth}
    \begin{minipage}{\textwidth}
        \begin{forest} for tree={
    grow=east,
    growth parent anchor=west,
    parent anchor=east,
    child anchor=west,
    anchor=parent,
    edge path={\noexpand\path[\forestoption{edge},draw] 
         (!u.parent anchor) -- +(4pt,0pt) |-  (.child anchor) 
         \forestoption{edge label};}
    }
        [Attention-based reasoning, xnode3, l sep=2mm,
            [Ask Me Anything\protect~\cite{Wu2016AskSources};\,MSMLAN\protect~\cite{Yu2019Multi-sourceAnswering};\,QANet\protect~\cite{Li2019VisualComprehension};\,KAT\protect~\cite{Gui2022KAT:Vision-and-Language};\,KAN\protect~\cite{Zhang2021RichAnswering};\,
            TRiG\protect~\cite{Gao2022Transform-Retrieve-Generate:Answering};\,\protect~\cite{Cao2022Knowledge-RoutedEmbedding};\,MMJG\protect~\cite{Wang2022Knowledge-EnhancedGuidance};\,\protect~\cite{Zhang2022QueryReasoning};\,VLC-BERT\protect~\cite{Ravi2023VLC-BERT:Knowledge};\\\protect~\cite{Song2023AnsweringPurpose};\,\protect~\cite{Wu2024ResolvingRetrieval},tnode3, l sep=16mm]]
        \end{forest}
        \vspace{-3mm}
    \end{minipage}\\
    \begin{minipage}{\textwidth}
        \begin{forest} for tree={
    grow=east,
    growth parent anchor=west,
    s sep=1mm,
    parent anchor=east,
    child anchor=west,
    anchor=parent,
    edge path={\noexpand\path[\forestoption{edge},draw] 
         (!u.parent anchor) -- +(6pt,0pt) |-  (.child anchor) 
         \forestoption{edge label};}
    }   
        [Structural reasoning, xnode3, l sep=4mm,
        [Interpretable reasoning, xnode31, l sep=2mm,
          [Att-MFH\protect~\cite{Riquelme2020ExplainingBase};\,KGENVQA\protect~\cite{Cao2024KnowledgeVQA};\,GRN\protect~\cite{Wang2024LetAnswering};\,SIRI\protect~\cite{Wang2023TowardsAnswering};\,\textcolor{brown}{MuKCoT}\protect~\cite{Qiu2024ExplainableAnswering},tnode4, l sep=4mm]],
          [Chain-of-Thought, xnode31, l sep=2mm,
            [\textcolor{brown}{IPVR}\protect~\cite{Chen2024VisualReasoning};\,\textcolor{brown}{MuKCoT}\protect~\cite{Qiu2024ExplainableAnswering};\,\textcolor{brown}{MCUD}\protect~\cite{Hu2024AVQA};\,\textcolor{brown}{\protect~\cite{Barezi2024DisentanglingKB-VQA}},tnode4, l sep=8mm]],
         [Graph-based reasoning, xnode31, l sep=2mm,
            [OotB\protect~\cite{Narasimhan2018OutAnswering};\,\protect~\cite{Singh2019FromReason};\,KG-AUG\protect~\cite{Li2020BoostingAggregation};\,GRUC\protect~\cite{Yu2020Cross-modalAnswering};\,Mucko\protect~\cite{Zhu2020Mucko:Answering};\,\protect~\cite{Zhang2022QueryReasoning};\,
            ConceptBert\protect~\cite{Ziaeefard2020TowardsAnswering};\,KI-Net\protect~\cite{Zhang2021ExplicitReasoning};\\Hypergraph Transformer\protect~\cite{Heo2022HypergraphAnswering};\,VQA-GNN\protect~\cite{Wang2023VQA-GNN:Answering};\,
            HCNMN\protect~\cite{Heo2022HypergraphAnswering};\, HyperMR\protect~\cite{Wang2024HyperMR:Answering},tnode4, l sep=8mm]]]
        \end{forest}
        \vspace{-3mm}
    \end{minipage}\\
    \begin{minipage}{\textwidth}
        \begin{forest} for tree={
    grow=east,
    growth parent anchor=west,
    parent anchor=east,
    child anchor=west,
    anchor=parent,
    edge path={\noexpand\path[\forestoption{edge},draw] 
         (!u.parent anchor) -- +(4pt,0pt) |-  (.child anchor) 
         \forestoption{edge label};}
    }
        [Dynamic reasoning, xnode3, l sep=2mm,
            [MAVEx\protect~\cite{Wu2022Multi-ModalVQA};\,DMMGR\protect~\cite{Li2022DynamicAnswering};\,MM-Reasoner\protect~\cite{Khademi2023MM-Reasoner:Answering};\,\protect~\cite{Xie2024Knowledge-AugmentedExplanation};\,SIRI\protect~\cite{Wang2023TowardsAnswering};\,
            LV$^2$-Net\protect~\cite{Wang2023TowardsAnswering};\,UniRVQA\protect~\cite{Deng2025EnablingAnswering};\,MIRAGE\protect~\cite{Wei2026MIRAGE:Chains},tnode3, l sep=16mm]]
        \end{forest}
    \end{minipage}
    \vspace{-4mm}
    \caption*{\footnotesize(c) Knowledge Reasoning}
    \end{minipage}

     \caption{Taxonomy of KB-VQA models, categorized into three key components: \revise{knowledge representation}, knowledge retrieval, and knowledge reasoning. Relevant discussions are in Sections III, IV, and V. Model abbreviations follow their original papers. Methods highlighted by \colorbox{brown}{ } are apply with LLM (implicit knowledge) only, and others are with access to explicit knowledge.}
    \label{fig:lit_surv}
\end{figure*}

\section{Knowledge Representation}\label{representation}
\revise{A fundamental challenge in KB-VQA is the effective semantic understanding of multimodal information. KB-VQA systems operate over diverse inputs, including image–question pairs, textual knowledge, and structured representations such as multimodal KBs and heterogeneous graphs. Learning appropriate knowledge representations is therefore crucial for accurate downstream knowledge retrieval and reasoning. Accordingly, this section organizes existing methods along two parallel stages in KB-VQA systems: query construction and knowledge modeling.} We begin with query construction \textcolor{blue}{(\cref{query})}, summarizing different methods of understanding image-question pairs that initiate the KB-VQA process. Query construction focuses on extracting and emphasizing salient information from both images and questions for downstream tasks. It then moves to knowledge modeling, which examines methods for fine-grained knowledge alignment \textcolor{blue}{(\cref{know align})}, with a particular focus on integrating graph-based knowledge structures \textcolor{blue}{(\cref{graph rep})} into KB-VQA. Finally, we summarize strategies for explicitly enhancing knowledge representation quality \textcolor{blue}{(\cref{know enhance})} to filter out noise and improve the expressiveness of knowledge.

\subsection{Query Construction}\label{query}
\revise{As shown in ~\cref{guideline}, query construction is the initial step in the KB-VQA pipeline, where multimodal information from the image and question are aligned to form a query representation that captures their combined semantic information. This query can be instantiated as a textual query, a dense multimodal embedding, or a structured graph, which serves as the direct input to downstream knowledge retrieval and reasoning.} Based on how visual information is processed, we categorize existing query construction methods into three approaches: image-to-text, image-to-embedding, and image-to-graph. These methods aim to capture fine-grained details and relationships rather than relying solely on global information. Image-to-text methods are relatively simple and enable direct access to powerful language models such as LLMs, but an inappropriate conversion design can lead to the loss of critical visual details. Image-to-embedding provides an alternative by encoding images directly into a high-dimensional semantic space, which is also the most dominant method in multimodal general area. Image-to-graph methods preserve spatial and semantic relationships, facilitating explicit and structural reasoning, though they often rely on predefined structures. \revise{Table I in the supplementary material presents a summary of image processing techniques applied by the KB-VQA methods included in this subsection.}

\subsubsection{Image-to-text}\label{i2t} 
\revise{While image-to-text transformation is a fundamental vision task, in KB-VQA it primarily serves as the interface for knowledge grounding. The image verbalization essentially reformulates the VQA task into text-based QA where reasoning happens in the language space. Since many external KBs are formed by textual documents, text features become a naturally compatible form for effective knowledge retrieval and downstream reasoning.} The image-to-text conversion for KB-VQA can be traced back to pre-LLM era, when \cite{Wu2016AskSources} developed a pipeline based on a multi-label CNN and LSTM network to generate dual representations: attribute-based features for structured query matching and caption-based features for reasoning. This method is later complemented by a question-guided knowledge selection scheme~\cite{Wu2018ImageKnowledge}. Such verbalization method continues in the space of KB-VQA after LLM comes into play~\cite{Gui2022KAT:Vision-and-Language,
Lin2022REVIVE:Answering,Gao2022Transform-Retrieve-Generate:Answering}. For example, TRiG~\cite{Gao2022Transform-Retrieve-Generate:Answering} converts visual information into textual representations by leveraging image captioning, dense object detection and Optical Character Recognition (OCR). REVIVE~\cite{Lin2022REVIVE:Answering} emphasizes the understanding of regional visual information and inter-region relationships using region detection. The model encodes spatial relationships between regions through position vectors, combining them with regional tags for LLM reasoning. Similarly, MAVEx~\cite{Wu2022Multi-ModalVQA} also explicitly leverages regional features for KB-VQA, but takes a different approach of internalizing object regions. MAVEx first generates candidate answers using a pre-trained VLM and extracts noun phrases through constituency parsing, which are then linked to detected object regions for generating the corresponding textual features. 

\textbf{Prompt Engineering.}\label{prompt}
In particular, image-to-text transformation serves as a common stage in prompt-based methods to bridge images and powerful LLMs~\cite{Yang2022AnVQA, Tiong2022Plug-and-PlayTraining, Guo2023FromModels}. PICa~\cite{Yang2022AnVQA} is the first that explores using LLMs to complete KB-VQA task by feeding multiple Question Answering (QA) pairs into the language model for in-context learning. Each QA pair is translated from the VQA and consists of image caption, question and ground truth answer. Img2LLM~\cite{Guo2023FromModels} utilizes a vision-language model and a question-generation model to translate the image content into synthetic questions verbally, which are fed to the LLM as part of the prompt. 
However, \cite{Hu2023PromptCap:GPT-3} argues that generic image captions can often lose important visual clues to answer. \revise{In response, PromptCap~\cite{Hu2023PromptCap:GPT-3} introduces a controllable captioning model.} The model is trained on synthetic data generated by GPT-3, enabling selective description of specific visual entities in the generated captions. Holding the same argument as~\cite{Hu2023PromptCap:GPT-3}, LAMOC~\cite{Du2023Zero-shotFeedback} introduces a reinforcement learning approach to optimize image captioning model performance through LLM-generated feedback. The model employs dual reward signals ( relevance and confidence scores) to rank generated captions and guide more task-aware caption generation, therefore achieving leading zero-shot performance in KB-VQA. Alternatively,~\cite{Wang2023FillingQuestions} closes the gap by prompting the LLM to generate new questions for the missing image information required to answer the original question with another filter to summarize generated questions for condensed useful information. In contrast, CBM~\cite{Salaberria2023ImageAnswering} verifies an effective image caption can facilitate the small model to achieve competitive results as GPT-3 without images. 

\subsubsection{Image-to-embedding}\label{i2c}
Encoding images to continuous embeddings is a standard process in multimodal tasks and is widely applied in KB-VQA, often achieved through Vision-Language Pretraining (VLP) models. General VLPs, such as CLIP~\cite{Radford2021LearningSupervision} and ALIGN~\cite{Jia2021ScalingSupervision}, are trained on vast image-text datasets to align visual and textual representations. These models typically extract global features from images. However, as fine-grain visual details are necessary for KB-VQA, which may not captured by pre-trained models. There are visual encoders specialized for KB-VQA tasks. For example, FLMR~\cite{Lin2023Fine-grainedAnswering} introduces and pretrains ViT~\cite{Dosovitskiy2020AnScale} to encode global and regional visual information. \revise{Similarly, ~\cite{Nguyen2022Coarse-to-FineAnswering} learns coarse-to-fine image representations to benefit the reasoning on different level of visual information. GeReA~\cite{Ma2024GeReA:Answering} adopts GradCAM~\cite{Selvaraju2017Grad-CAM:Localization} to find the most relevant regions and then encode visual regions into embeddings for the process of MLLMs.}

\subsubsection{Image-to-graph}\label{i2g}
Several works have explored structuring image content into graphs that encode objects, attributes, and relationships. Such a graph is commonly referred as a scene graph or visual graph. These structured representations can serve as an semantic abstraction of the image content that facilitates reasoning and knowledge retrieval in KB-VQA. For example, Mucko~\cite{Zhu2020Mucko:Answering} and GRUC~\cite{Yu2020Cross-modalAnswering} utilize DenseCap~\cite{Johnson2016DenseCap:Captioning} to generate multiple captions about the question. \revise{Then the semantic graph is constructed by a semantic graph parsing model~\cite{Anderson2016SPICE:Evaluation}. Each node represents the name
or attribute of an object extracted from the captions and the edge represents the relationship between object nodes.} VQA-GNN~\cite{Wang2023VQA-GNN:Answering} applies a pre-trained scene graph generation model~\cite{Tang2020UnbiasedTraining} and an object detection model~\cite{Zhang2021VinVL:Models} to jointly formulate the scene graph to represent structured image context. However,~\cite{Heo2022HypergraphAnswering} challenges the previous work cannot capture high-order semantics for multi-hop answering. Therefore, they propose to construct hypergraphs on both question and image, where each hyperedge will connect an arbitrary number of nodes and has partial order itself.

\subsection{\revise{Knowledge Modeling}}\label{know rep}
\noindent While query construction captures essential information from image-question input, effectively representing knowledge in KB-VQA further presents unique challenges: \revise{\textit{Knowledge Alignment} addresses fine-grained semantic mismatches when grounding and aligning heterogeneous sources, such as structured KGs and unstructured textual descriptions referring to the same entity. \textit{Graph-based Knowledge Representation} enables structured and interpretable reasoning but demands specialized integration mechanisms. \textit{Knowledge Enhancement} is needed to filter and refine the large-scale and noisy external knowledge to prevent error propagation in reasoning.}

\subsubsection{Knowledge Alignment}\label{know align}
We divide knowledge alignment into two main types: \textbf{cross-modal alignment} and \textbf{intra-modal alignment}. Cross-modal alignment aims to establish connections of features across different modalities (e.g. linking images to text descriptions), which has been widely researched in the field of multimodal learning~\cite{Jabeen2023ALearning, Wu2023MultimodalSurvey}. However, KB-VQA usually requires the alignment to be performed with a finer granularity, as the salient features might be highly localized in both image and text modalities. On the other hand, intra-modal alignment harmonizes the semantic inconsistency between knowledge and queries within the same modality.

The earlier strategy for cross-modal alignment involves converting multimodal knowledge into descriptive sentences. For example, both LaKo~\cite{Chen2022LaKo:Injection} and VLC-BERT\cite{Ravi2023VLC-BERT:Knowledge} convert KG triples into descriptive sentences for integration. LaKo uniquely delays knowledge-query attention interaction until the decoding phase, allowing independent knowledge integration before multimodal fusion. However, this process often distorts the original structure and semantic information, as observed by~\cite{Song2021KVL-BERT:Reasoning} when incorporating ConceptNet~\cite{Speer2017ConceptNetKnowledge}. To mitigate this, KVL-BERT~\cite{Song2021KVL-BERT:Reasoning} introduces relative position embeddings and a masked self-attention mechanism, which restricts token visibility and provides more structured guidance for external knowledge utilization.

Another branch of cross-modal alignment approaches uses specialized mapping networks to bridge the semantic gap between KBs and multimodal queries~\cite{Lin2023Fine-grainedAnswering,Feng2024Implicit-Knowledge-GuidedKB-VQA, Li2024CognitiveAlignment}. Specifically, FLMR~\cite{Lin2023Fine-grainedAnswering} trains a two-layer Multilayer Perceptron (MLP) to project visual and text features into a unified semantic space in the context of knowledge retrieval. The relevance between each query-document pair is calculated on token-level to achieve fine-grained understanding. Additionally,
\cite{Li2024CognitiveAlignment} present a Visual Knowledge Aligner, which consists of a cross-attention and a self-attention layer before MLPs, to distill important visual knowledge from a smaller autoregressive language model~\cite{Zhang2022OPT:Models}. 
 
On the other hand, \cite{Chae2022Uncertainty-basedBase} further highlights potential semantic inconsistencies between KBs and pre-trained feature extraction models. Therefore, it is necessary to align the semantic details even the information are presented in the same modalities. Correspondingly, a novel semantic consistency model is introduced, based on S-BERT sentence embedding method~\cite{Reimers2019Sentence-BERT:BERT-Networks} and the knowledge to be used will adaptively adjusted by the uncertainty measurement. Meanwhile, \cite{Salemi2023AAnswering} also highlights the challenge to encode documents and queries into a shared embedding space as the encoder for each is asymmetric. Therefore, DEDR~\cite{Salemi2023AAnswering} is proposed as an iterative knowledge distillation method, where the uni-modal and \revise{multimodal} encoder will iteratively teach each other to minimize the Kullback–Leibler divergence between their embedding space. Meanwhile,~\cite{Hu2024AVQA} observed that the biased training sample distribution could lead to limited generalizability in unseen domains. Therefore, they proposed a multimodal unbiased contrastive distillation method, which calculates the entropy-adjusted answer frequencies on the training set to label each original sample as biased or unbiased sample.

\subsubsection{Graph-based Knowledge Representation}\label{graph rep}
Graph-based knowledge representations serve as another fundamental format in KB-VQA. A common practice involves extracting knowledge from KBs and structuring it as a graph to facilitate traceable reasoning~\cite{Narasimhan2018OutAnswering,Ziaeefard2020TowardsAnswering, Marino2021KRISP:VQA}. Each node stores semantic representations of knowledge entities, while edges capture the relationships between them to formulate knowledge triplets: \texttt{(head, relation, tail)}. KRISP~\cite{Marino2021KRISP:VQA} is one of the earliest works focusing on constructing KGs for KB-VQA. Specifically, it extracts knowledge concepts from DBpedia~\cite{Auer2007DBpedia:Data},ConceptNet~\cite{Speer2017ConceptNetKnowledge}, VisualGenome~\cite{Krishna2017VisualAnnotations}, and hasPartKB~\cite{Bhakthavatsalam2020DoRelations} to cover knowledge from commonsense to spatial relationships. The recognized concepts will then be processed and stored on each node of the Relational Graph Convolutional Network (RGCN)~\cite{Schlichtkrull2018ModelingNetworks}. 

However, knowledge described by textual concepts lacks visual information to assist cross-modal understanding. MuKEA~\cite{Ding2022MuKEA:Answering} addresses this issue by learning to build vision-aware KGs. It extracts multimodal triplets using pre-trained LXMERT~\cite{Tan2019LXMERT:Transformers} with the relevance affinity matrix to process image-question pairs, where the head entity contains visual content, the tail entity represents the answer conditioned on image-question pair, and their high-order relation is learned. The triplet representation is learnable using the combination of TransE, triplet consistency, and semantic consistency losses to unify multimodal semantic space. Later, Wang et al.~\cite{Wang2024Cross-modalityAnswering} argue that MuKEA has the limitation for its inability to model multiple relations between question-related visual objects. To address this, they propose CMRL, which extends single relations to multiple relations in the aforementioned KG triplets. CMRL introduces a global-local loss mechanism for representation learning: the global loss learns cross-modality relations by contrasting positive and negative multimodal triplets while linking ground-truth answer entities. The local loss maintains semantic diversity among these relations to prevent them from converging to a single representation.

On the other hand, there is effort to enhance the overall graph representation from diverse focus. For example, REAVL~\cite{Rao2023Retrieval-basedPre-training} propose a Link Prediction module, making the model use the KG structure jointly with the corresponding \revise{multimodal} data to reason about missing links in KG. Additionally, Chen et al.~\cite{Chen2023DeconfoundedInference} investigate into the bias problem in existing KB-VQA systems that tend to learn highly co-occurring object relationships,  while overlooking casual ones. They propose KECVQG, which uses an Attribute Factorization Module to separate causal from non-causal attributes, ultimately debiasing the knowledge representation in the graph. OWKRL~\cite{Zheng2024KnowledgeReasoning} further investigates how the system is capable of realizing
``unknown" samples and then continuing to update incrementally without retraining when adding new samples. Specifically, a novel graph-based self-cross transformer and a continual learning method are proposed to represent the knowledge and keep the open-world representation updated. 

\textbf{Advanced Graph Structures.} \revise{In addition to regular graph-based methods, several works have explored alternative graph structures with the goal of capturing higher-order semantics and relationships.} For instance,~\cite{Heo2022HypergraphAnswering} introduces the concept of hypergraphs and proposes transformer-based attention mechanism over hypergraphs to encode and reason high-order associations between objects. Zhang et at.~\cite{Zhang2023TowardKnowledge} argue that the existing work of modeling knowledge is lack of capability of distinguishing different granularity of concepts. HCNMN is presented with a Hierarchy Concept Graph, in which the more general the concept, the higher the layer in hierarchy. Two types of edges are utilized to connect the concepts in the same or different granularity layer, which are intra-layer edge and inter-layer edge.
Alternatively,~\cite{Wang2024HyperMR:Answering} attempts to model the hierarchical structure of knowledge by constructing a hyperbolic space based on Hyperbolic GCN~\cite{Chami2019HyperbolicNetworks}. Compared to Euclidean space, Bourgain's theorem~\cite{Linial1995TheApplications} shows that the hyperbolic space is better to obtain lower distortion for three structures therefore more suitable for modeling the knowledge. 

\textbf{Key-value Memory Network}
To handle multiple knowledge facts expanding from questions,~\cite{Su2018LearningAnswering} opts for Key-value Memory Network~\cite{Miller2016Key-ValueDocuments} to save graph knowledge as key-value pairs, where the entity is set as key and the subject and relation are set as value. Such a memory structure is flexible enough to reason about the object, the subject and relation of a knowledge triplet. Based on this, DMMGR~\cite{Li2022DynamicAnswering} is proposed so that the key-value memory and retrieved subgraphs can be used to update with each other iteratively. Along with this dynamic design, the issue of involving irrelevant key can be alleviated. Additionally, KM$^4$~\cite{Zheng2021KM4:Modulation} emphasizes the importance of visual information and introduces it into memory-based modeling framework. Through the proposed mutual modulation, the knowledge-based key-value memory network can then reason over image and question jointly and iteratively.

\subsubsection{Knowledge Enhancement} \label{know enhance}
Another persistent challenge lies with KB-VQA task is the noisy external knowledge. Although various knowledge retrieval methods have been proposed to filter out irrelevant information from KBs, proactively enhancing the overall quality of the KBs themselves could further benefit retrieval efficiency. 

\textbf{Explicit Knowledge Enhancement.} The first and an intuitive solution is to condense the KB into a less noisy one. For example, \cite{Hao2024KnowledgeVQA} proposes condensing knowledge into two levels: first, the passages are distilled into concepts using MLLMs; second, the LLM is employed to summarize and extract the essential knowledge from the passages. Olano et al.~\cite{Adjali2023ExplicitEntities} first suggest constructing a reduced KB from Wikipedia by identifying 2,400 unique entities. They then created a knowledge-enhanced passage representation by integrating structural entity knowledge from the KG. While such entity-centric knowledge reduction method may be less flexible for application across diverse KBs, the enhanced passage representation offers valuable insights for improving subsequent retrieval performance. Additionally,~\cite{Adjali2023ExplicitEntities} still working on entity-centric knowledge representation enhancement, where the entity span is recognized by a pre-trained Wikipedia2Vec~\cite{Yamada2016JointDisambiguation} and implicitly injected, where the entity and words sharing the embedding space.~\cite{Sun2023BreakingVQA} introduces another method of knowledge representation enhancement, which is the generation of dynamic knowledge representation. By learning through a fully connected layer, the acquired semantic knowledge representation can be transformed into a dynamic vector that VLP models can understand.

\textbf{Implicit Knowledge Enhancement.} In addition to enhancing explicit knowledge for improved quality, there is a line of work aims at strengthening the implicit knowledge within models. For instance, SelTDA~\cite{Khan2023Q:Images} introduces a novel data augmentation strategy to fine-tune the VLP for specialized knowledge. Meanwhile, they highlight that datasets for KB-VQA in non-natural image domains are significantly smaller than those for general-purpose VQA. Specifically, SelTDA leverages the VLM and the target dataset to build a teacher model that generates question-answer pseudo-labels directly from images. This approach enables pseudo-labeling of unlabeled images, allowing the model to fine-tune the initial VLM on unseen visual knowledge using the augmented dataset. PLLMKI~\cite{Hu2024PromptingAnswering} proposes a cross-LLM knowledge enhancement approach. The first LLM is required to generate background knowledge given a set of questions. Then the generated knowledge will be fed as auxiliary information to enhance the inference of the second LLM. Despite their effectiveness, these methods do not directly correct implicit knowledge errors in LLMs. While solutions such as knowledge editing techniques~\cite{Wang2025KnowledgeSurvey} or continual learning~\cite{Wang2024AApplication} have been widely explored for general LLMs, how to effectively apply them to KB-VQA remains an open challenge.

\section{Knowledge Retrieval} \label{knowledge retreival section}
Knowledge retrieval is a foundational component of KB-VQA systems to get access to external KBs. \revise{It determines the critical external knowledge that will be leveraged by the downstream reasoning modules. Unlike generic retrieval methods~\cite{Gao2023Retrieval-AugmentedSurvey,Mei2025AGeneration}, where retrieval primarily serves to supply contextually relevant text, KB-VQA must bridge the semantic gap by aligning image details with both textual questions and specific entities in KGs or encyclopedic entries. Consequently, retrieval in KB-VQA is typically a more challenging and granular search compared to general RAG.} Various retrieval methods can be classified based on different knowledge formats: Document retrieval \textcolor{blue}{(\cref{passage retr})} focuses on extracting relevant unstructured documents by measuring the semantic similarity between the question and textual knowledge sources. This section will also cover the studies that leverage unstructured \revise{multimodal} KBs. Graph retrieval approaches \textcolor{blue}{(\cref{graph retr})} leverage structured KGs by exploring entity-relation connections to identify relevant knowledge paths. Additionally, recent work has introduced retrieval post-processing techniques \textcolor{blue}{(\cref{ret post})} that refine and enhance the initially retrieved knowledge through methods like re-ranking or iterative retrieval. Each of these retrieval approaches offers unique advantages in accessing and utilizing knowledge.

\subsection{Document Retrieval}\label{passage retr}
\noindent To retrieve the most relevant documents from large corpora, Dense Passage Retrieval (DPR)~\cite{Karpukhin2020DenseAnswering} is one of fundamental approaches in KB-VQA systems. \revise{DPR maps queries and passages into a high-dimensional dense vector space, which outperforms traditional keyword-based sparse retrieval. By leveraging bi-encoders, it transforms queries ($q$) and documents ($d$) into continuous embeddings that capture latent semantic relationships. Relevance $R(q, d)$ is then efficiently computed as the cosine similarity of these dense vectors, as detailed in \textcolor{blue}{\cref{passage retreival score function}}. More recent work has explored more sophisticated scoring functions that leverage fine-grained interactions between queries and knowledge. For instance, MAVEx~\cite{Wu2022Multi-ModalVQA} implements BertScore~\cite{Zhang2019BERTScore:BERT}, which performs greedy matching between query and document tokens. The token similarities are then aggregated using inverse document frequency weights. ColBERT~\cite{Khattab2020ColBERT} represents the successful evolution of this trend, which has also been adopted by KB-VQA systems~\cite{Luo2021Weakly-SupervisedAnswering}. Specifically, the proposed ``MaxSim" operator only looks for the best match for each query token in the document, which significantly alleviate the high computation cost of BertScore. This is efficient enough to search millions of documents, hence adopted by some strong KB-VQA methods~\cite{Luo2021Weakly-SupervisedAnswering,Lin2023Fine-grainedAnswering}.} FLMR~\cite{Lin2023Fine-grainedAnswering} further extends this mechanism to multimodal contexts by employing a simple and lightly trained MLP, making it particularly well-suited for getting access to fine-grain knowledge. Based on FLMR, PreFLMR~\cite{Lin2024PreFLMR:Retrievers} investigates the scaling laws in the late-interaction retrieval. 
\begin{table}[t]
    \centering
    \caption{\revise{Popular scoring functions for measuring query-document relevance in knowledge retrieval. }}
    \begin{tabular}{c|c}
    \hline
    \textbf{Relevance Score}  & \textbf{Formula} \\
    \hline
     Cosine Similarity    & $R(q, d)=\frac{\mathbf{q}^\mathsf{T}\cdot \mathbf{d}}{||\mathbf{q}||\cdot||\mathbf{d}||}$ \\
     \hline
     BertScore            &$R(q, d)=\frac{\sum_{q_i\in q}\text{idf}(q_i)\text{max}_{d_j\in d}\mathbf{q_i}^\mathsf{T}\mathbf{d_j}}{\sum_{q_i\in q}\text{idf}(q_i)}$\\
     \hline
     Late Interaction     &$R(q, d)=\sum_{q_i\in q}\text{max}_{d_j\in d}\mathbf{q_i}^\mathsf{T}\mathbf{d_j}$\\
    \hline
    \end{tabular}
    \label{passage retreival score function}
\end{table}

\textbf{Training Strategy.} \revise{The primary objective in training retrievers is to establish a shared embedding space where both knowledge and queries are mapped parametrically. \cref{tab:training_strat} summarizes the main training strategies for effective retrievers. It is noted that contrastive learning~\cite{Khosla2020SupervisedLearning} is widely employed, which optimizes the retriever by maximizing the similarity of positive knowledge-query pairs while minimizing the similarity of negative pairs~\cite{Karpukhin2020DenseAnswering, Chen2022MuRAG:Text,Lin2023Fine-grainedAnswering, Deng2025EnablingAnswering}.
The general contrastive loss function is:}
\begin{equation} 
\mathcal{L}_{contrast}=-\sum_{q, d^+} \log \frac{\exp(R(q, d^+)}{\exp(R(q, d^+)) + \sum_{d^+\in D_n} \exp(R(q, d^+)}
\end{equation}
\revise{where $R(q,d^+)$ denotes the relevance score between query input $q$ and positive knowledge $d^+$, whose calculation has been detailed in \cref{passage retreival score function}.}
During contrastive learning, determining the ground-truth positive knowledge $d^+$ is a key challenge. A common practice is the use of pseudo labels, where knowledge snippets containing ground-truth answers are treated as relevant knowledge~\cite{Luo2021Weakly-SupervisedAnswering, Lin2023Fine-grainedAnswering}. However, such label assignment can mistakenly classify irrelevant knowledge that coincidentally contains the answer as relevant, thereby introducing noise into the model. Therefore, RA-VQA~\cite{Lin2022RetrievalKnowledge} proposes a novel label assignment method that jointly incorporates feedback from the generator. In this approach, ground-truth knowledge must both contain the answer and enable the generator to produce the correct answer. Alternatively, CEIK~\cite{Wang2024CoordinatingVQA} aims to coordinate the conflicts between implicit and explicit knowledge during the training process and a piece of knowledge can be judged as positive when the prediction output using this knowledge is correct. Further increasing efforts also emphasize the importance of joint training between the retriever and generator. UnifER~\cite{Guo2022AVQA} explores an inspiring joint training framework that the retriever can benefit from the updates of the generator, although its performance falls short compared to current leading models. Similarly, \cite{Wu2022Multi-ModalVQA} proposes a joint framework for a large pretraining KB-VQA model, where the retriever and generator partially share parametric representations to enhance their mutual understanding. UniRVQA~\cite{Deng2025EnablingAnswering} extends the joint training with a unified framework and a self-reflections task to achieve leading performance.
\revise{More recently, \cite{Hong2025Knowledge-basedFiltering} propose VLM-PRF, where the knowledge retriever is optimized using reinforcement learning. By employing answer accuracy and format consistency as reward signals, the RL framework supervises the model’s ability to dynamically invoke tools and filter irrelevant retrieved noise, which ensures precise reasoning and structural adherence.}

\begin{table}[b]
\centering
\caption{\revise{Main training strategies for retrievers in KB-VQA systems. CL: Contrastive Learning; RL: Reinforcement Learning; KD: Knowledge Distillation; JT: Joint Training.}}
\begin{tabularx}{\columnwidth}{@{}
  >{\hsize=0.3\hsize\arraybackslash}X
  >{\hsize=1.9\hsize\raggedright\arraybackslash}X
  >{\hsize=0.8\hsize\raggedright\arraybackslash}X
@{}}
\toprule
\textbf{Strategy} & \textbf{Training Mechanism}& \textbf{Key methods} \\ \midrule
\textbf{CL} & Maximizes similarity for relevant (positive) pairs and minimizing it for negatives. & RA-VQA~\cite{Lin2022RetrievalKnowledge}, CEIK~\cite{Wang2024CoordinatingVQA} \\  \addlinespace
\textbf{RL} & Uses reward signals to train retrieval agent. & VLM-PRF~\cite{Hong2025Knowledge-basedFiltering} \\ \addlinespace
\textbf{KD} & Trains a student model to learn from a powerful teacher LLM outputs. &  DEDR~\cite{Salemi2023AAnswering}\\ 
\addlinespace
\textbf{JT} & Optimizes retriever and generator simultaneously through shared parameters or feedback loops. & UnifER~\cite{Guo2022AVQA}, MAVEx~\cite{Wu2022Multi-ModalVQA}, UniRVQA~\cite{Deng2025EnablingAnswering} \\\bottomrule
\label{tab:training_strat}
\end{tabularx}
\end{table}

\textbf{Multimodal Retrieval.} As a significant extension of text-based retrieval, multimodal retrieval broadens access to diverse knowledge sources in KB-VQA by incorporating both textual and visual data, ultimately enhancing overall performance.

MuRAG~\cite{Chen2022MuRAG:Text} is one of the pioneers to explore multimodal retrieval in KB-VQA. It proposes a retrieval-augmented transformer, which is pretrained with a mixture of large-scale image-text and text-only corpora. On the other hand, both~\cite{Reichman2023Cross-ModalAnswering} and~\cite{Lerner2023MultimodalAnswering} introduce the multimodal Inverse Cloze Task to train the multimodal retriever. ICT~\cite{Lee2019LatentAnswering} optimizes a model to match the sentence to its greater context with a portion of context masked, which enables the model to understand which pieces of information are closely related and which are more distantly related.~\cite{Reichman2023Cross-ModalAnswering} breaks down the ICT task into four sub-tasks: bidirectional matching between image and textual document as well as bidirectional matching between titles and image-passage pairs. While~\cite{Lerner2023MultimodalAnswering} adapts the ICT task to 
learn the modality-invariant representation of the named entities, the intuition being to ``replace" the text-named entities learned by base model BERT~\cite{Devlin2019BERT:Understanding} with the ``visual" entities present in the images.~\cite{Salemi2023AAnswering} highlights that the property of passage retrieval is dealing with asymmetric input modalities: the query information is multimodal while the information items (passages) are uni-modal. To close the gap, DEDR~\cite{Salemi2023AAnswering} reframes the task into a symmetric bi-encoder architecture through grouping each textual document with an empty picture as the input. Through the iterative knowledge distillation, the document encoder and the query encoder will teach each other and close the modality gap. This method improves the retrieval performance of asymmetric uni-modal retrieval that can be extended to multimodal retrieval task. More recently, GeMKR~\cite{Long2024GenerativeModels} creatively translates the retrieval task into a multimodal generative task. The method simplifies the knowledge retrieval by directly generating identifiers of relevant documents for queries rather than retrieving from a large-scale corpus.~\cite{Lerner2024Cross-ModalAnswering} exclusively focus on the entity-centric cross-modal retrieval, where they highlights the importance to align various visual representations of the identical entity with textual descriptions. Therefore, they propose to combine mono-modal retrieval with cross-modal retrieval to enhance the multimodal retrieval performance. \rev{BayesRAG~\cite{Li2026BayesRAG:Generation} leverages Bayesian inference and evidence fusion to model cross-modal consistency, enabling more reliable and coherent multimodal retrieval beyond similarity-based methods.}

\subsection{Graph Retrieval}\label{graph retr}
\noindent For KG retrieval, many studies utilize context graphs constructed from the image-question context. These approaches analyze semantic relationships between nodes, using them as queries for knowledge retrieval~\cite{Narasimhan2018StraightAnswering,Li2020BoostingAggregation, Zheng2021KnowledgeModel}. For instance,~\cite{Narasimhan2018StraightAnswering} formulates the retrieval task as a standard multinomial classification task. Specifically, the embedded words of the question will be fed into a LSTM to predict the relations between entities from a fix relation set. Then all knowledge triplets in KG containing the predicted relations will be kept for answer inference. Such learning-based methods exhibit satisfying robustness. However, they are largely depend on ground-truth knowledge facts which serve as extra supervised information and may fail
when there are no available ground-truth knowledge facts or when external knowledge are less necessary for the referred questions. 

Therefore, KG-Aug~\cite{Li2020BoostingAggregation} proposes to retrieve a context-aware knowledge subgraph. The method is based on the anchor entities extracted from the question and the name of prominent visual objects. Expanding from the anchor entity to their first-order neighbors, the large-scare KGs can be reduced to a context-aware subgraph eventually. The proposed aggregation method to reason over two subgraph can be found in \textcolor{blue}{\cref{reasoning}}. Similarly, VQA-GNN~\cite{Wang2023VQA-GNN:Answering} constructs the concept subgraph by the entities extracted from images and questions with relevant concept matching. The proposed module will build new knowledge triplets if a local concept entity from image is found adjacent to a retrieved entity from KGs, which allows better comprehension of contextual concepts. 

You et al.~\cite{You2023AAnswering} highlight that conventional methods for detecting objects in images often overlook scene entity information, leading to semantic ambiguity. To reduce the ambiguity, RR-VEL~\cite{You2023AAnswering} emphasizes on recognizing key entities in images to replace question referents. In the retrieval stage, knowledge triplets are encoded using BERT~\cite{Devlin2019BERT:Understanding} without retraining graph structures, in contrast to standard approaches~\cite{Zhu2020Mucko:Answering, Garderes2020ConceptBert:Answering}, as the authors argue that retaining these structures hinders the efficient utilization of external knowledge. While most of work focus on exploring high-order relationships between knowledge and questions, DSAMR~\cite{Sun2024DSAMR:Answering} highlights the negative effect of unnecessary knowledge and proposes a hyper-graph extraction method to extract the optimal knowledge instead of noisy knowledge for the answer predictions. The hyper-graph extraction module employs the relevance affinity matrix as guidance, which describes the relevance of each knowledge hyperedge feature to the question hyperedge feature for ranking. The selected knowledge hyperedge feature will then be weighted by the Gumbel-Softmax.

\subsection{Retrieval Post-processing}\label{ret post}
\noindent While much of the existing literature discussed above emphasizes one-way retrieval methods aimed at reducing noise from large KBs, it is acknowledged that not all retrieval results are satisfactory. Even documents deemed relevant can still introduce noise into the system. As a result, two key areas of research have emerged: \textbf{post-processing}, which refines and reorganizes retrieved knowledge, and \textbf{dynamic retrieval}, where iterative and adaptive retrieval mechanisms leverage the results of initial searches to inform and refine subsequent retrievals. Both approaches seek to refine the retrieval process and enhance the system's robustness.

\textbf{Post-processing Strategies.} The post-processing methods that refine the retrieved knowledge can be broadly divided into three ways: re-ranking~\cite{Wu2018ImageKnowledge} , filtering~\cite{Zhou2023PromptingVQA}, and consolidation~\cite{Wang2023FillingQuestions, Liu2024LookGeneration}. For example,~\cite{Wu2018ImageKnowledge} implements a question-guided knowledge selection scheme after the retrieval process by evaluating the semantic similarity between question and retrieved knowledge. The scheme aims to exclude the knowledge irrelevant to the question even it is relevant to the image content. Zhou et al.~\cite{Zhou2023PromptingVQA} proposes a knowledge filter that classifies the selected knowledge into three classes (useful, harmful and neutral) so that the harmful knowledge will be filtered out. The classifier consists 11 features including knowledge confident, the contradiction between question and knowledge etc.. Both~\cite{Wang2023FillingQuestions} and~\cite{Liu2024LookGeneration} refine the knowledge for the in-context learning prompts. Specifically, ~\cite{Wang2023FillingQuestions} trains a summarizer and a filter, where the former summarizes the question and answer into a narrative description and the later will assess the helpfulness of the summary to the answer inference stage. In addition, LV$^2$-Net~\cite{Liu2024LookGeneration} injects the logic-verified knowledge refinement module, where the visual and answer are combined to refine the external knowledge information and facilitate the construction of logic-enhanced prompts.~\cite{Wang2024SoftVQA} adaptively select knowledge context by utilizing the relevance scores of each knowledge tokens produced by head. Lastly, the weighted knowledge representations of all heads
are concatenated together to serve as a soft knowledge
prompt for the answer generation. Inspired from GradCAM~\cite{Selvaraju2017Grad-CAM:Localization}, ~\cite{Li2024DiversifyVQA} proposes to evaluate the contribution of each token in the decision context rather than the raw attention scores. Based on that, a novel rationalization module, which consists of the automatic rationale generated by LLM and the mechanistic rational selected by token contributions, is proposed to further filter out irrelevant context. ~\cite{Xu2024Multi-ModalAnswering} propose the \revise{multimodal} validation module (MMV) to evaluate the confidence of each retrieved knowledge fact via images and questions, which preserves knowledge candidates effective for inferring answers. With the similar purpose, Boter~\cite{Hao2024Boter:VQA} utilize the initiated MLLM as the selector to evaluate the importance of knowledge and the filtered knowledge will be used to trained the answerer. In return, the updated answerer will be used to optimize the decision of selector.

\textbf{Dynamic Retrieval.} From another aspect, dynamic retrieval improves the retrieval process by adaptively selecting retrieval results or re-conducting the process until satisfactory results are achieved. For example,  Wiki-LLaVa~\cite{Caffagni2024Wiki-LLaVA:LLMs} proposes a hierarchical retrieval pipeline, where predominant documents are retrieved first, then the most relevant passages will be identified with the Contriver architecture~\cite{Izacard2021UnsupervisedLearning} for enhanced precision. Recently, there also has been a surge of interest in applying LLMs as autonomous agents to interact with external APIs, making dynamic decisions until achieve specific goals~\cite{Nakano2021WebGPT:Feedback,Liu2023LearningKnowledgeb}. For instance, AVIS~\cite{Hu2023AVIS:Agent} enables a tree-search decision-making that can dynamically activate image or web page retrieval for various knowledge augmentation or directly involve a LLM to answer the question. Another work~\cite{Wang2023TowardsAnswering} employs multiple agents at the different stage of KB-VQA - responder, seeker and integrator. Collaborating with the responser, the seeker can retrieve the implicit knowledge from LLMs to generate relevant issues and aggregate the information from responses.

\begin{figure}[t]
    \centering
   \centerline{\includegraphics{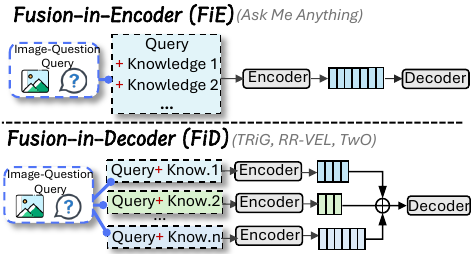}}
    \caption{\revise{Key frameworks of fusing queries and retrieved knowledge in KB-VQA methods. FiE concatenates queries with retrieved knowledge and processes them within the encoder. FiD encodes each query-knowledge pair independently and aggregates the resulting representations in the decoder. Representative methods are listed in gray.}}
    \label{fusion}
\end{figure}
\section{Knowledge reasoning}\label{reasoning}
\begin{figure*}[t]
    \centering  \includegraphics[width=\textwidth]{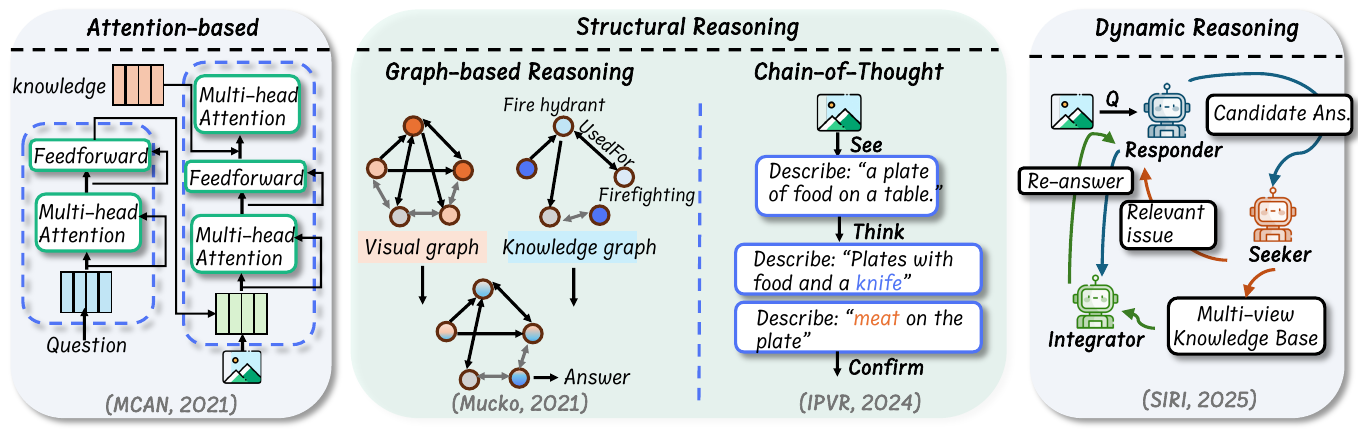}
    \caption{\revise{Illustration of major knowledge reasoning paradigms, including attention-based reasoning, graph-based structural reasoning, CoT reasoning, and dynamic agent-based reasoning. The figure provides conceptual visualizations to clarify complex reasoning mechanisms rather than detailed implementations; readers are referred to the original papers for model specifics.}}
    \label{reasoning_type}
\end{figure*}

\noindent In KB-VQA, reasoning is the inevitable final stage. A strong reasoning model can logically combine image-question information with the retrieved knowledge and wisely infer answers from context. 
%This section will explore the existing methods, which are broadly categorized into attention-based reasoning, structural reasoning, and dynamic reasoning. 
\revise{This section reviews existing reasoning approaches, which are categorized according to their primary reasoning carrier and methodological contribution.}
Attention-based reasoning 
\textcolor{blue}{(\cref{attention reasoning})} \revise{comprises methods whose core contribution lies in the design of specialized attention mechanisms for implicit information integration, enabling the model to selectively emphasize relevant details for precise reasoning.} Structural reasoning consists of graph-based approaches \textcolor{blue}{(\cref{graph reasoning})} and LLM-based Chain-of-Thought (CoT) methods \textcolor{blue}{(\cref{cot reasoning})} \revise{, where reasoning is guided by explicit structures such as relational graphs or sequential logic thereby naturally improving interpretability} \textcolor{blue}{(\cref{interpretable reasoning})}. Dynamic reasoning \textcolor{blue}{(\cref{dynmaic reasoining})} focuses on adaptive refinement of the inference process and utilizes intermediate signals for more accurate responses. \revise{A comparative summary of the reasoning paradigms is provided in \cref{reasoning_paradigm_table}, and representative implementations of each paradigm are illustrated in \revise{\cref{reasoning_type}}.}

\subsection{Attention-based reasoning} \label{attention reasoning}
% \noindent Attention mechanism~\cite{Vaswani2017AttentionNeed} is a cornerstone in a wide range of reasoning tasks. Its success in cross-modal reasoning tasks stems from the ability to model intricate dependencies across multimodal information, adaptively highlight the important elements of inputs while amplifying task-critical signals for precise answer inference.

% Earlier extractive methods relied on selecting answers based on attention scores over knowledge snippets~\cite{Yu2019Multi-sourceAnswering}. Recent approaches often leverage attention-based generative language models to synthesize answers by reasoning over integrated contexts and evidence. A critical component in this process is the fusion of the query with the retrieved knowledge, which can occur either in the encoding or the decoding stage, as shown in Figure\textcolor{red}{~\ref{fusion}}. Fusion-in-Encoder (FiE) combines the query and all evidence early by concatenating them into a single input for the encoder, enabling immediate interaction between the inputs but potentially struggling with numerous knowledge~\cite{Wu2016AskSources}. In contrast, Fusion-in-Decoder (FiD)~\cite{Izacard2021LeveragingAnswering} processes each passage separately in the encoder and defers fusion until the decoding stage, allowing the decoder to selectively attend to relevant details and better handle large evidence sets~\cite{ Gao2022Transform-Retrieve-Generate:Answering, Izacard2021LeveragingAnswering,You2023AAnswering,Si2023ComboVQA}. 

% During the fusion and reasoning stage, several works develop tailored attention module to enhance intra-modal interactions and focus on salient input regions. These approaches can be broadly categorized into \textit{fusion-based attention, adaptive attention}, and \textit{attention augmentation}. 
\begin{table}[b]
\centering
\caption{\revise{Summary of reasoning paradigms in KB-VQA}}
\label{tab:reasoning_paradigms}
\begin{tabularx}{\columnwidth}{@{}
  >{\hsize=0.67\hsize\raggedright\arraybackslash}X
  >{\hsize=1.18\hsize\raggedright\arraybackslash}X
  >{\hsize=1.15\hsize\raggedright\arraybackslash}X
@{}}
\toprule
\textbf{Reasoning Paradigm} & 
\textbf{Main Reasoning Mechanism} & 
\textbf{Research Focus} \\
\midrule
\revise{Attention-based} &
\revise{Implicit relevance modeling across multimodal representations} &

\revise{Specialized attention module and cross modal interaction design.} \\
\midrule
\revise{Structural reasoning (graph-based)} &
\revise{Explicit relational modeling via predefined graph structures} &
\revise{Advanced graph structure design for graph-based reasoning.} \\
\midrule
\revise{Structural reasoning (CoT based)} &
\revise{Explicit step-by-step logical decomposition} & 
\revise{Tasks decomposition and interpretable reasoning process.}
 \\
\midrule
\revise{Dynamic reasoning} &
\revise{Adaptive refinement and control of the reasoning process} &
\revise{Iterative reasoning control, self-refinement, feedback-driven reasoning} \\
\bottomrule
\label{reasoning_paradigm_table}
\end{tabularx}
\end{table}

\revise{Attention mechanism~\cite{Vaswani2017AttentionNeed} serves as the computational backbone for reasoning in KB-VQA. The primary challenge lies in modeling the dependencies between the question ($q$), visual features ($v$), and retrieved knowledge snippets ($k$). 
These methods focus on the architectural integration of heterogeneous features. KAT~\cite{Gui2022KAT:Vision-and-Language} utilizes a \textit{Knowledge-to-Vision} cross-attention layer. Given visual features $f_v$ and a concatenated knowledge representation $f_k$ (comprising implicit and explicit knowledge), the interaction is:
\begin{equation}
    f_{out} = \text{Softmax}\left(\frac{(f_v W^Q)(f_k W^K)^T}{\sqrt{d}}\right)(f_k W^V)
\end{equation}
Other variants like QANet~\cite{Li2019VisualComprehension} use dual attention blocks for context-to-question interaction, while Cao et al.~\cite{Cao2022Knowledge-RoutedEmbedding} utilize knowledge embeddings as $K$ and $V$ conditioned on image features to route the attention. MAVEx~\cite{Wu2022Multi-ModalVQA} extends this via parallel attention modules for multi-granular aggregation.}

\revise{Recent approaches leverage Transformer-based generative models to synthesize answers by reasoning over integrated contexts. The choice of fusion strategies greatly impacts how retrieved evidence interacts with the query (\cref{fusion}).}

\revise{\textbf{Fusion-in-Encoder (FiE):} In FiE, the input sequence is formed by concatenating the question with all $N$ retrieved passages: $X = [q; k_1; k_2; \dots; k_N]$. The self-attention operation allows for an all-to-all interaction:
\begin{equation}
    \text{Attention}(Q, K, V) = \text{softmax}\left(\frac{XW^Q (XW^K)^T}{\sqrt{d_k}}\right) XW^V
\end{equation}
While powerful, the $O(L^2)$ complexity (where $L$ is the total sequence length) makes it computationally prohibitive for handling the ``numerous knowledge" problem often found in large-scale retrieval~\cite{Wu2016AskSources}.}

\revise{\textbf{Fusion-in-Decoder (FiD):} To alleviate this, FiD~\cite{Izacard2021LeveragingAnswering} encodes each pair $(q, k_i)$ independently. If $E_i$ is the encoding of the $i$-th passage, the decoder performs cross-attention over the union of all encoded representations:
\begin{equation}
    H_{dec} = \text{CrossAttention}(h_{query}, [E_1; \dots; E_N], [E_1; \dots; E_N])
\end{equation}
This allows the model to scale to hundreds of passages, as the decoder selectively attends to relevant details within the large evidence set~\cite{Gao2022Transform-Retrieve-Generate:Answering, You2023AAnswering, Si2023ComboVQA}.}

\revise{\textbf{Adaptive Attention.} The adaptive attention methods instead focus on dynamic selection of salient information. To prevent noise from irrelevant knowledge, KAN~\cite{Zhang2021RichAnswering} introduces a gating mechanism. The final reasoning feature $F$ is a weighted combination of visual attention $a_v$ and knowledge attention $a_k$:
\begin{equation}
    F = g \cdot a_v + (1-g) \cdot a_k
\end{equation}
where $g = \sigma(W_g[a_v; a_k])$ is a learnable sigmoid gate. Similarly, MMJG~\cite{Wang2022Knowledge-EnhancedGuidance} and Wu et al.~\cite{Wu2024ResolvingRetrieval} use scoring strategies to dynamically weight relevant entities and relations in the KG.}

\revise{\textbf{Attention Augmentation.} \cite{Zhang2022QueryReasoning} proposes to augment the attention vector $\alpha$ with a memory-guided component $m$:
\begin{equation}
    \alpha_{aug} = \text{Softmax}(\text{MLP}(h_t, m_{t-1}))
\end{equation}
The model can therefore maintain consistency across visual and knowledge queries by generating parallel attention vectors augmented with memorized intermediate results.}

\subsection{Structural Reasoning}
Structural reasoning methods represent another reasoning approach with regularized processes and explicit relational modeling, which naturally align with human-like problem-solving that breaks down complex tasks into interpretable steps. In this section, we discuss two types of structural reasoning: \textit{graph-based reasoning} and \textit{CoT}. Graph-based reasoning focuses on navigating through nodes and edges in KGs to derive answers, while confined by limited knowledge access. CoT prompting guides LLMs through sequential reasoning steps to articulate intermediate logic and conclude the answer. 
\subsubsection{Graph-based Reasoning} \label{graph reasoning}
To jointly consider all entities and their relationships, one of the prevalent techniques for graph-based reasoning is graph-based neural networks. For example,~\cite{Narasimhan2018OutAnswering} adopts GCN~\cite{Kipf2016Semi-SupervisedNetworks} to reason about the final answer by neighborhood aggregation and message passing mechanism. To further reason over the multi-relational graph, \cite{Singh2019FromReason} proposes to adopt Gated Graph Neural Network~\cite{Li2015GatedNetworks} for sequential outputs. While KG-Aug~\cite{Li2020BoostingAggregation} proposes to incorporated the structured knowledge into the conventional VQA approach, as described in the last section, through a context-aware deep fusion mechanism, where the embedded knowledge is distilled through the constructed auxiliary features and fused together with visual-question embeds. KRISP~\cite{Marino2021KRISP:VQA} processes the structured knowledge by RGCN~\cite{Schlichtkrull2018ModelingNetworks} individually and then combines it with the implicit vector generated from BERT by the dot product. The combination is named as the symbolic vectors. Then the answer will be selected with the highest scoring answer from both answer vectors. Additionally, ~\cite{Li2022DynamicAnswering} conduct max-pooling over the graph at the reasoning stage with the spatial-aware attention results.

\textbf{Cross-modal Reasoning.} Yu et al.~\cite{Yu2020Cross-modalAnswering} argue that the entities on the concept graph provide insufficient knowledge to reason about the globally optimal answer and need to be complemented with correlated knowledge from other modalities. Therefore, they proposes a cross-modal reasoning method, which is a memory-based architecture by sequencing a recurrent Graph-based Read, Update and Control module. Meanwhile, Zhu et al. also propose Mucko~\cite{Zhu2020Mucko:Answering} to address the same question. To gather the complementary information from visual graph and semantic graph, the cross-modal convolution is adopted, including visual-to-fact convolution and semantic-to-fact convolution. Additionally, KI-Net~\cite{Zhang2021ExplicitReasoning} explicitly connects KGs to the visual graph through the proposed knowledge incorporation network, where the topological extension will ensure the candidate entities in the scene graph are directly related to the visual observation and remain semantically consistent with corresponding classes in the KG. Alternatively,~\cite{Ziaeefard2020TowardsAnswering} applies Compact Trilinear Interaction~\cite{Do2019CompactAnswering} to fuse question embeddings, scene
graph visual features, and concept features into a single vector for joint representation. 

However, VQA-GNN~\cite{Wang2023VQA-GNN:Answering} argues that methods that only perform a unidirectional fusion from unstructured knowledge to structured knowledge, limiting their potential to capture joint reasoning over the heterogeneous modalities of knowledge. Therefore, a new bidirectional multimodal GNN technique is introduced to perform inter-modal message
passing for reasoning that mitigates representational gaps between modalities. In addition, ~\cite{Zhang2022QueryReasoning} argues that most reasoning methods that utilize neural modules networks~\cite{Andreas2016NeuralNetworks} cannot effectively generate queries and tend to loosely integrate the knowledge in reasoning process. To address this limitation, a novel query and attention augmentation method is proposed, which tightly couples knowledge incorporation with reasoning.

% \textbf{Hypergraph-based Reasoning.} Another key focus of graph-based reasoning is to enable inference over complex graph structures. For example, ~\cite{Heo2022HypergraphAnswering}introduces hypergraphs to learn high-order semantics of a question and a knowledge base. In addition, they applies the guided-attention~\cite{Tsai2019MultimodalSequences} blocks to learn the correlations between knowledge hyperedges and question hyperedges by inter-attention mechanism.  Zhang et al.~\cite{Zhang2023TowardKnowledge} also propose a hierarchical reasoning process based on the hierarchy concept graph they initiate. The essence of it is to model the propagation of knowledge across different levels of granularity and reasoning steps by a novel attention mechanism that operates on both the inter-layer and intra-layer edges of the hierarchical graph.Additionally,~\cite{Wang2024HyperMR:Answering} performs hyperbolic reasoning over Poincare Ball Transformation. Taking advantage of the property that hyperbolic space can be warped and adjusted, HyperMR~\cite{Wang2024HyperMR:Answering} divides the reasoning process into different stages so as to find the most
% suitable spatial structure and conduct curvature assignments for each stage.

\textbf{Hypergraph-based Reasoning.} \revise{Another key focus in KB-VQA research involves enabling sophisticated inference over non-linear and high-dimensional graph structures. To capture higher-order relational dependencies, Heo et al. \cite{Heo2022HypergraphAnswering} represent both the question and the KB as hypergraphs $\mathcal{H} = \{\mathcal{V}, \mathcal{E}\}$. In this formulation, a hyperedge $h \in \mathcal{E}$ connects an arbitrary number of nodes to encapsulate $n$-ary relations and high-order semantics without the constraints of fixed edge lengths. 
The proposed Hypergraph Transformer constructs these edges via $n$-hop graph walks, where the tail node of a preceding walk serves as the head node for the subsequent step. To learn cross-modal associations, the model utilizes guided-attention blocks~\cite{Tsai2019MultimodalSequences} to compute the inter-association between knowledge hyperedges $E^k$ and question hyperedges $E^q$. This framework mitigates the common over-smoothing problem found in standard graph convolutional networks by updating representations through hyperedge matching rather than iterative message passing.}

\revise{Addressing the requirement for multi-granular reasoning, Zhang et al. \cite{Zhang2023TowardKnowledge} propose a hierarchical structured reasoning across varying levels of abstraction. This method employs a specialized attention mechanism that operates on both inter-layer edges (vertical taxonomic relations) and intra-layer edges (horizontal relations), modeling the propagation of knowledge through a multi-level reasoning chain.}

\revise{Furthermore, to better represent the power-law distributions and tree-like hierarchies inherent in large KBs, HyperMR \cite{Wang2024HyperMR:Answering} performs hyperbolic reasoning within the Poincaré ball model $\mathbb{D}^n = \{ \mathbf{x} \in \mathbb{R}^n : \|\mathbf{x}\| < 1 \}$. By leveraging Möbius addition ($\oplus_c$) and stage-specific curvature assignments, the model dynamically adjusts the spatial structure to fit the hierarchical density of the required knowledge.} \revise{A comparative illustration of representative regular graph-based and hypergraph-based reasoning model is provided in Supplementary Figure 1.}

\subsubsection{Chain-of-Thought}\label{cot reasoning}
Alternatively, CoT focuses on leveraging the reasoning capabilities of LLMs. In particular, CoT prompting, which uses cues like ``Let’s think step by step" or in-context examples with rationales, guides LLMs to generate intermediate reasoning steps. The key is designing effective prompts to help LLMs decompose complex questions and arrive at correct answers.

Chen et al. \cite{Chen2024VisualReasoning} introduce a step-by-step reasoning framework that guides LLMs to ``see" (analyze the visual context), ``think" (generate textual descriptions of visual contents), and ``confirm" (validate the reasoning process through rationale generation). This structured approach ensures a thorough understanding across modalities during CoT. MuKCoT~\cite{Qiu2024ExplainableAnswering} generates auto-labeling reasoning chains via LLMs prompting and distills smaller Vision-and-language models from which to perform CoT reasoning for \revise{KB-VQA} tasks.~\cite{Barezi2024DisentanglingKB-VQA} proposes to first decompose the questions into simpler sub-questions with the LLM and then determine whether the external knowledge is needed when prompting the LLMs.
Furthermore,  MCUD~\cite{Hu2024AVQA} explores the hierarchy CoT reasoning method that sequentially locates the contextual sentences in the knowledge corpus, generates the rationale, and infers the answer.

\subsubsection{Interpretable Reasoning.}\label{interpretable reasoning}
In KB-VQA reasoning stage, interpretability is essential for helping researchers understand how neural networks arrive at the final answers. Such explainable reasoning process is typically built upon KGs or LLMs, where the former offer structured pathways for inference~\cite{Zhang2021ExplicitReasoning} and the later can utilize carefully designed prompts~\cite{Chen2024VisualReasoning} to generate explicit rationales which guide the generation, making the answering process transparent.

Early in ~\cite{Riquelme2020ExplainingBase}, a textual explanation module is proposed by combining the internal representations of different component of the model, including attended image feature, the original question feature and the answer embedding. Then the module is trained with a softmax cross-entropy loss on the explanation dataset~\cite{Park2018MultimodalEvidence}. GRN~\cite{Wang2024LetAnswering} conducts step-by-step graph-based entity routing to predict the answer. At each step, GRN keeps an entity
score vector to show how likely of each entity to be activated
as the answer, and a transition matrix representing the transition probability from one entity to another. KGENVQA~\cite{Cao2024KnowledgeVQA} directly lets the LLM to generate the knowledge to support its answer and demonstrate the reasoning process. Their experiments show that incorporating explicit knowledge statements covering different aspects of the image-question pair may increase the chance of getting the correct answer. Similarly, SIRI~\cite{Wang2023TowardsAnswering} and MuKCoT~\cite{Qiu2024ExplainableAnswering} also utilize LLMs to generate the rationale of each decision so that the top-down reasoning process will be explained. 

\subsection{Dynamic Reasoning}\label{dynmaic reasoining}
\noindent Dynamic reasoning represents a fundamental shift from end-to-end KB-VQA approaches, which follow a one-pass reasoning process by generating a single answer without revisiting reasoning process thus limiting flexibility and accuracy. In contrast, dynamic reasoning involves an iterative or adaptive method, where the model refines its reasoning progressively therefore enhances the robustness of the system.

For example, DMMGR~\cite{Li2022DynamicAnswering} will iteratively conduct reasoning process for $T$ steps. At each step, the model identifies the most relevant knowledge triplet, retrieves the corresponding semantic values of the nodes, and applies spatial-aware image graph attention for reasoning. MM-Reasoner~\cite{Khademi2023MM-Reasoner:Answering} will first prompt an LLM to extract knowledge and uses the potential answers predicted by the VLM to iteratively update and optimize the prompt, refining its answer. Xie et al.~\cite{Xie2024Knowledge-AugmentedExplanation} further emphasize that the generated answers and explanations of the context should be consistent in semantics. Therefore, they propose an iterative consensus generator which consists of the rough answer generation as well as the iterative answer and explanation generation. The answers and explanations will be iteratively refined to achieve enhanced consistency. MAVEx~\cite{Wu2022Multi-ModalVQA} proposes an answer validation module, where the answer can be mostly support by retrieved knowledge will be validated as the final answer. UniRVQA~\cite{Deng2009ImageNet:Database} proposes self-reflective mechanism that enable the MLLM to evaluate its implicit knowledge boundary and then adaptively choose to answer with or without explicit knowledge.

More recently, agent-based methods have gained popularity in the design of dynamic reasoning frameworks. For instance,~\cite{Wang2023TowardsAnswering} adopts multiple agents that can initiate the re-answer process based on each other's responses and agents will vote among the answer pool for the final answer. LV$^2$-Net~\cite{Liu2024LookGeneration} also proposes to use the logic to enhance the reasoning process by checking the logical structure among image-question pair, primary answer, knowledge, ground-truth answers twice in the whole KB-VQA procedure. \rev{MIRAGE~\cite{Wei2026MIRAGE:Chains} introduces a test-time scalable, multi-chain reasoning framework that performs parallel, graph-structured retrieval and cross-chain verification to improve the answering performance.}
\begin{table*}[t]
    \centering
    \caption{Comparison of popular KB-VQA datasets. \revise{+ symbols denote a relative ordinal scale indicating the level of capability required to answer questions in each dataset, as inferred from the comparative performance analyses reported in the original papers. Additional + symbols indicate a higher difficulty or stronger emphasis on the corresponding ability.}}
    \setlength{\tabcolsep}{4pt}        % Adjust column spacing
    % \begin{threeparttable}
        \begin{tabular}{l c c >{\raggedright\arraybackslash}p{1.8cm} | c | c | >{\raggedright\arraybackslash}p{8.7cm} }
        \toprule
            \multirowcell{2}[-2pt][l]{\textbf{Dataset}\\\textbf{(Year)}} & \multirowcell{2}[-2pt][l]{\textbf{\# Img}} & \multirowcell{2}[-2pt][l]{\textbf{\# Ques}} & \multirowcell{2}[-2pt][l]{\textbf{Knowledge}\\\textbf{Base}} & \multicolumn{2}{c|}{\textbf{Focus}} & \multirowcell{2}[-2pt][l]{\textbf{Challenges}} \\
        \cmidrule(lr){5-6}
            &  &  &  & \textbf{Reason} & \textbf{Retrieval} & \\
        \midrule
            \makecell[tl]{KB-VQA~\cite{Wang2017ExplicitAnswering} \\ (2015)} & 700 & 2402 & DBpedia & \textbf{+} & \textbf{+} & Commonsense KB-VQA.  \\
        \midrule  
            \makecell[tl]{FVQA~\cite{Wang2018FVQA:Answering}\\(2018)}  & 2190 & 5826 & DBpedia/ ConceptNet/ WebChild &\textbf{++} &\textbf{+} & Providing supporting rationales and multiple knowledge sources for question reasoning. \\
       \midrule     
            \makecell[tl]{ GQA~\cite{Hudson2019GQA:Answering}\\(2018)}  & 113K & 22M & Visual Genome & \textbf{+++} & \textbf{+} & Reasoning over large-scale KGs. \\
        \midrule    
            \makecell[tl]{KVQA~\cite{Shah2019KVQA:Answering}\\(2019)}  & 24K & 183K  & Wikidata & \textbf{++} & \textbf{++} & The first dataset for the world knowledge-aware VQA task. Reasoning over large-scale KGs and recognizing named entities. \\
        \midrule    
            \makecell[tl]{ OK-VQA~\cite{Marino2019OK-VQA:Knowledge}\\(2021)}  & 14,031 & 14,055  & Wikipedia/ GoogleSearch & \textbf{++} & \textbf{+++} & ``Open-world" KB-VQA benchmark. Focus on knowledge retrieval and reasoning at the same time. \\
        \midrule
            \makecell[tl]{ S3VQA~\cite{Jain2021SelectAnswering}\\(2021)}  & 6,765 & 6,765  & Wikipedia/ GoogleSearch & \textbf{+++} & \textbf{+++} & Interpretable reasoning process. \\
        \midrule    
            \makecell[tl]{ A-OKVQA~\cite{Schwenk2022A-OKVQA:Knowledge}\\(2022)} & 23,692 & 24,903  & Wikipedia/ GoogleSearch & \textbf{+++} & \textbf{+++} & Complex reasoning questions with supporting rationale annotations.\\
        \midrule    
            \makecell[tl]{ViQUAE~\cite{Lerner2022ViQuAEEntities}\\(2022)}  & ~3,700 & ~3,300 & Wikipedia & \textbf{++} & \textbf{++++} & Recognizing named entities for \revise{KB-VQA}. Multimodal Retrieval. \\
        \midrule    
            \makecell[tl]{InfoSeek~\cite{Chen2023CanQuestions}\\(2023)} & - & 1.35M  & \makecell[tl]{Wikipedia\\Wikidata} & \textbf{++} & \textbf{++++} & Recognizing named entities for \revise{KB-VQA}. Questions require fine-grained retrieval.\\
        \midrule    
            \makecell[tl]{E-VQA~\cite{Mensink2023EncyclopedicCategories}\\(2023)}  & 514K &  221K & WIT~\cite{Srinivasan2021WIT:Learning} & \textbf{+++} & \textbf{++++} & Fine-grained instance QA involving complex multi-hop reasoning. \\
        \midrule    
            \makecell[tl]{SnapNTell~\cite{Qiu2024SnapNTell:LLM}\\(2024)} & 75,680 & 75,680  & Wikipedia/ Wikidata & \textbf{+++} & \textbf{++++} & Asks questions about fine-grained instances. Questions are complicated without knowledge hints, requiring deep reasoning. \\
        \bottomrule
        \end{tabular}
        
    % \end{threeparttable}
    
    \label{dataset_summary}
\end{table*}

\begin{table}[t]
    \centering
    \caption{\revise{Summary of leading methods across OK-VQA, InfoSeek, E-VQA, GQA, and FVQA datasets. The corresponding evaluation metrics are indicated in parentheses, whose details can be found in \cref{sec:eval}.}}
    \setlength{\tabcolsep}{6pt}
    \begin{tabular}{l l l c}
    \toprule
    \textbf{Method} & \textbf{Base Model} & \textbf{External KB} & \textbf{Score} \\
    \midrule
    \rowcolor[gray]{0.95} \multicolumn{4}{c}{\textbf{OK-VQA} (VQAScore)} \\
    MAVEx~\cite{Wu2022Multi-ModalVQA} & VisualBERT & ConceptNet & 38.4 \\
    REVIVE~\cite{Lin2022REVIVE:Answering} & GPT-3 & Wikipedia & 58.0 \\
    PromptCap~\cite{Hu2023PromptCap:GPT-3} & OFA & GPT-3 & 60.4 \\
    RA-VQA-v2~\cite{Lin2023Fine-grainedAnswering} & BLIP2-$\mathrm{T5_{XL}}$ & GoogleSearch & 60.8 \\
    UniRVQA~\cite{Deng2025EnablingAnswering} & InstructBLIP-$\mathrm{T5_{XL}}$ & GoogleSearch & 61.6 \\
    MuKCoT~\cite{Qiu2024ExplainableAnswering} & ChatGPT-3.5 & - & 70.7 \\
    \midrule
    \rowcolor[gray]{0.95} \multicolumn{4}{c}{\textbf{InfoSeek} (VQA Accuracy)} \\
    Wiki-LLaVa~\cite{Caffagni2024Wiki-LLaVA:LLMs} & LLaMA-3.1-8B & Wikipedia & 27.1 \\
    MMKB-RAG~\cite{Ling2025MMKB-RAG:Framework} & Qwen-2-7B & Wikipedia & 36.4 \\
    VLM-PRF~\cite{Hong2025Knowledge-basedFiltering} & Qwen-2.5VL-7B & Wikipedia & 42.8 \\       ReflectiVA~\cite{Cocchi2025AugmentingAnswering} & LLaMA-3.1-8B & Wikipedia & 40.1 \\
    PreFLMR~\cite{Lin2023Fine-grainedAnswering} & ViT-H/ColBERTv2 & Wikipedia & 48.6 \\
    \midrule
    \rowcolor[gray]{0.95} \multicolumn{4}{c}{\textbf{E-VQA} (BEM)} \\
    Wiki-LLaVa~\cite{Caffagni2024Wiki-LLaVA:LLMs} & LLaMA-3.1-8B &  WIT & 19.6 \\
    ReflectiVA~\cite{Cocchi2025AugmentingAnswering} & LLaMA-3.1-8B &  WIT & 35.5 \\
    MMKB-RAG~\cite{Ling2025MMKB-RAG:Framework} & Qwen-2-7B &  WIT & 35.9 \\
    VLM-PRF~\cite{Hong2025Knowledge-basedFiltering} & InternVL3-8B &  WIT & 39.2 \\
    PreFLMR~\cite{Lin2024PreFLMR:Retrievers} & ViT-L/ColBERTv2 &  WIT & 45.0 \\
    \midrule
    \rowcolor[gray]{0.95} \multicolumn{4}{c}{\textbf{GQA} (VQAScore)} \\
    InstructBLIP~\cite{Dai2023InstructBLIP:Tuning} & Vicuna-13B & - & 49.5\\
    CFM~\cite{Nguyen2022Coarse-to-FineAnswering} & - & Visual Genome & 73.6  \\
    VQA-GNN~\cite{Wang2023VQA-GNN:Answering} & RoBERTa & Visual Genome & 90.3  \\
    \midrule
    \rowcolor[gray]{0.95} \multicolumn{4}{c}{\textbf{A-OKVQA} (VQAScore)} \\
    KRISP~\cite{Marino2021KRISP:VQA} & BERT & - & 27.1 \\
    Prophet~\cite{Yu2025Prophet:Answering} & MCAN-large & GPT-3 & 58.5 \\
    PromptCap~\cite{Hu2023PromptCap:GPT-3} & OFA & GPT-3 & 59.6 \\
    InstructBLIP~\cite{Dai2023InstructBLIP:Tuning} & Vicuna-7B & - & 62.1\\
    \bottomrule
    \end{tabular}
    \label{tab:vertical_summary}
\end{table}

\section{Benchmarks and Evaluation}\label{benchmarks}
% benchmark summarize here 
\subsection{Datasets}
\noindent Mainstream datasets in KB-VQA differ in their evaluation focus on different on different system capabilities, making a systematic comparison and clear differentiation essential. As summarized in \revise{\cref{dataset_summary}}, these datasets are primarily designed to validate two key capabilities of KB-VQA systems: (1) accurate \textbf{reasoning} over intensive knowledge, and (2) effective  \textbf{retrieving} the relevant expertise or encyclopedic knowledge.

The earliest KB-VQA datasets are KB-VQA~\cite{Wang2017ExplicitAnswering} and FVQA~\cite{Wang2018FVQA:Answering}. KB-VQA~\cite{Wang2017ExplicitAnswering} selects 700 images from MS COCO~\cite{Lin2014MicrosoftContext}, with questions supported by the DBpedia KG~\cite{Auer2007DBpedia:Data}. FVQA~\cite{Wang2018FVQA:Answering} extends KB-VQA~\cite{Wang2017ExplicitAnswering} by incorporating more types of images from Microsoft COCO and ImageNet\cite{Deng2009ImageNet:Database} with multiple KBs supporting (DBpedia, ConceptNet\cite{Speer2017ConceptNetKnowledge}, and WebChild~\cite{Tandon2014WebChild}). It also annotates rationales to aid reasoning. While these datasets ensure external knowledge is necessary, they rely on ``closed" knowledge by constructing questions from a fixed KG. As a result, once the relevant fact is retrieved, answering becomes trivial, requiring minimal reasoning. Moreover, their knowledge scope is strictly limited to predefined sources, restricting the evaluation of generalizable knowledge retrieval and reasoning. 

KVQA~\cite{Shah2019KVQA:Answering} introduced a different approach by utilizing images from Wikidata~\cite{Vrandecic2014Wikidata}. The dataset primarily focused on recognizing specific named entities (e.g., \texttt{Barack Obama}) and retrieving factual knowledge from Wikipedia. Being different from earlier datasets, KVQA prioritizes bridging images with world knowledge rather than commonsense retrieval or reasoning. On the other hand, the concurrent work GQA~\cite{Hudson2019GQA:Answering} is designed to test compositional reasoning in VQA, leveraging structured annotations from Visual Genome~\cite{Krishna2017VisualAnnotations} about images from COCO~\cite{Lin2014MicrosoftContext} and Flickr~\cite{Thomee2016YFCC100M}. GQA requires reasoning over object relations and attributes but focuses less on integrating large-scale external knowledge.

One of the most influential KB-VQA datasets is OK-VQA~\cite{Marino2019OK-VQA:Knowledge}. OK-VQA provides more challenging knowledge-based questions and does not constrain knowledge to a predefined source, making it the first ``open-world" KB-VQA benchmark. However, OK-VQA still has limitation in favoring simple factual lookups (e.g. \texttt{What is the capital of this country?}) rather than complex reasoning. S3VQA~\cite{Jain2021SelectAnswering},  built on OK-VQA, modifies questions to better evaluate semantic understanding but still focuses on single retrieval step. A-OKVQA~\cite{Schwenk2022A-OKVQA:Knowledge} enhances OK-VQA by increasing question complexity and adding rationale annotations, enabling models to be evaluated on reasoning process.

More recent datasets push fine-grained retrieval from both structured and unstructured sources. ViQUAE~\cite{Lerner2022ViQuAEEntities} focuses on multimodal entity alignment. InfoSeek~\cite{Chen2023CanQuestions}, based on OVEN~\cite{Hu2023Open-domainEntities}, emphasizes fine-grained entity recognition. E(ncyclopedic)-VQA~\cite{Mensink2023EncyclopedicCategories} extends category diversities with challenging two-hop reasoning questions. SnapNTell~\cite{Qiu2024SnapNTell:LLM} further complicates questions by removing knowledge hints about entities, requiring deeper reasoning skills (e.g. instead of \texttt{Is the object from the 50s or 60s?}, it asks \texttt{What is its current geographical status?}). These datasets move beyond simple retrieval or reasoning toward real-world knowledge-seeking behavior. However, their reliance on external retrieval introduces challenges in maintaining answer consistency across diverse sources.

\subsection{Evaluation and Performance Comparison}\label{sec:eval}
\noindent KB-VQA systems are evaluated mainly based on two aspect: answer generation and knowledge retrieval performance:

(1) \textit{Exact Match (EM)} measures the exact matching between the generated answer $\hat{a}$ and the answer set $S$, where $\#s(\hat{a})$ is the occurrences of $\hat{a}$ in $S$: $\mathrm{EM}(\hat{a}, S)=\mathrm{min}(\#s(\hat{a}),1)$

(2) \textit{VQAScore~\cite{Agrawal2017VQA:Answering}} is generally considered an improved evaluation metric over \textit{EM} on datasets where multiple correct answers are provided for a single question (e.g. OK-VQA~\cite{Marino2019OK-VQA:Knowledge}). It partially rewards the model if it generates a less popular answer among human responses: 
$\mathrm{VQAScore}(\hat{a}, S)=\mathrm{min}(\#s(\hat{a})/3,1)
$

\revise{(3) \textit{BEM~\cite{Bulian2022TomaytoEvaluation}} is a semantic evaluation metric that utilizes a pre-trained BERT model to determine the relevance between a candidate answer and the ground truth. It maps both sequences into a high-dimensional embedding space to calculate a matching score, which allows for a more flexible assessment than exact string matching by capturing synonyms and semantic equivalence in open-ended KB-VQA tasks~\cite{Mensink2023EncyclopedicCategories}.}

(4) \textit{Recall (Recall@K)} is a prevalent evaluation metric for information retrieval systems, measuring the proportion of relevant documents successfully retrieved from the total $K$ returned relevant documents. For datasets such as InfoSeek~\cite{Chen2023CanQuestions}, which provide ground-truth relevant passages, recall serves as a crucial metric to assess how effectively the system retrieves the expected relevant knowledge.

(5)~\textit{Pseudo Relevance Recall (PRR@K):} In the more open-ended cases ~\cite{Lin2022RetrievalKnowledge, Lin2023Fine-grainedAnswering} where ground-truth knowledge is not specified, pseudo-relevance labels are usually adopted. The retrieval performance will be evaluated by counting the number of questions that successfully retrieve documents with correct answers contained in top-K retrieval results. However, it is worth noted that pseudo-relevance labels rely on direct keyword matching, they may introduce noise and inaccuracies, leading to potential biases in evaluation.

\revise{\cref{tab:vertical_summary} provides a performance comparison of representative KB-VQA methods on selected datasets, where clear performance differences of different methods across main KB-VQA benchmarks reflect their distinct knowledge and reasoning requirements. On OK-VQA and A-OKVQA, methods that explicitly leverage LLMs as implicit knowledge consistently outperform earlier vision-centric or retrieval-based approaches, which showcases the importance of integrating large-scale implicit knowledge and multi-step reasoning. In contrast, InfoSeek and E-VQA place stronger emphasis on fine-grained entity retrieval and grounding, where retrieval-augmented and multimodal retrieval methods achieve more competitive results, though overall performance remains lower, indicating the difficulty of precise knowledge access. For GQA, which relies on structured visual knowledge, models built on explicit scene graph representations and supervised reasoning achieve substantially higher scores, reflecting the benefits of well-defined visual knowledge structures. Overall, the comparison suggests that no single approach dominates across datasets. Instead, effective KB-VQA solutions must align their knowledge sources and reasoning mechanisms with dataset-specific characteristics.}

\section{\revise{Challenges and Future Directions}}\label{challenges}

\noindent\textbf{\revise{Ethical and Safety Considerations in KB-VQA.}} \revise{The deployment of KB-VQA systems in real-world raises important ethical and safety concerns. First, privacy risks may arise when systems retrieve or reason over sensitive content or external knowledge sources containing personal or proprietary information. Second, KB-VQA systems are vulnerable to misinformation propagation when retrieved knowledge is biased, potentially leading to factually incorrect or misleading answers. These risks are further amplified when LLMs are used as implicit knowledge sources. Third, biases present in visual data, KBs, or retrieval mechanisms may be inherited and reinforced during reasoning, resulting in unfair outputs. Nevertheless, systematically evaluating ethical risks and ensuring reliable and safe KB-VQA systems remain open challenges that warrant further investigation.}

\noindent\textbf{Unified Benchmark.}
While earlier KB-VQA datasets relied on closed KBs, recent benchmarks have shifted toward open-domain retrieval and complex reasoning with various focuses, leading to different and less fair comparisons. \rev{Additionally, the performance and efficiency of many LLM-based KB-VQA methods are largely determined by their backbone models, yet inference cost and latency are rarely reported in a standardized manner. Some prior works on agent-based and iterative RAG frameworks~\cite{Hu2023AVIS:Agent} also explicitly identify high latency as their key limitations. This further underscores the need for standardized evaluation of efficiency, which remains an open challenge and is critical for real-world deployment. }

\noindent\textbf{\revise{Hallucinations in KB-VQA.}}
\revise{As LLMs increasingly shape KB-VQA systems, hallucinations originating from their latent space remain underexplored. Although retrieval-augmented methods are widely adopted to ground answer generation in explicit evidence, hallucination is rarely addressed as a standalone research problem. Most existing approaches mitigate it implicitly through retrieval grounding or structured reasoning, without explicit detection or correction mechanisms. Developing hallucination-aware KB-VQA frameworks, including reliable detection, therefore remains significant.}~\rev{Future research should leverage specialized evaluation paradigms, such as~\cite{Moratelli2026BenchmarkingModels}, which is a benchmark that actively assesses models' ability to ``deflect" or remain silent when external knowledge is insufficient. Integrating such metrics will be crucial for developing KB-VQA systems that prioritize grounded factual consistency over speculative generation.}

\noindent\textbf{\revise{Challenges in Practical Deployment.}}
\revise{While KB-VQA architectures have advanced, specialized sectors present unique hurdles that prevent easy real-world deployment. Taking healthcare as an example, where medical images are often captured in special modalities, significant modifications to image processing modules are required to capture the key visual features for differential diagnosis. Furthermore, the ``black-box" nature of neural reasoning poses ethical risks, as medical practitioners require verifiable evidence and right-to-explanation for any AI-suggested diagnosis. To address these gaps, specialized research is emerging~\cite{Lin2023MedicalSurvey,Zi2025RSVLM-QA:Answering}, utilizing domain-specific designs to better align multimodal evidence. Future work should continue toward building specialized methods, as well as general-purpose KB-VQA techniques that can benefit downstream specialized applications.}

\noindent\textbf{Reasoning over heterogeneous KBs.} Existing methods often leverage a single knowledge source, which limits their ability to handle diverse inputs and effectively resolve conflicting information. Future research could therefore focus on multi-source knowledge fusion strategies that can integrate heterogeneous knowledge, along with robust mechanisms for conflict resolution to improve reliability and robustness.
%\section*{Acknowledgments}
%This should be a simple paragraph before the References to thank those individuals and institutions who have supported your work on this article.

\bibliographystyle{IEEEtran}
\bibliography{references,bib_control}
\section{Author Biography}
\vspace{-\baselineskip}
\vspace{-\baselineskip}
\begin{IEEEbiographynophoto}{Jiaqi Deng}
received her Bachelor’s degree from the University of Sydney. She is currently pursuing a Ph.D. degree at the University of Technology Sydney. Jiaqi's research interests include multimodal reasoning, knowledge-based visual question answering systems, and their practical translation to real-world applications.
\end{IEEEbiographynophoto}
\vspace{-\baselineskip}
\vspace{-\baselineskip}
\begin{IEEEbiographynophoto}{Zonghan Wu}
received a Ph.D. in computer science from the University of Technology Sydney, Australia. He is currently an assistant professor with the Faculty of Shanghai AI Finance School, East China Normal University. His research interests include graph machine learning and large language models.
\end{IEEEbiographynophoto}
\vspace{-\baselineskip}
\vspace{-\baselineskip}
\begin{IEEEbiographynophoto}{Huan Huo}
is currently an Associate Professor at the School of Computer Science in the Faculty of Engineering and Information Technology at the University of Technology Sydney working in the fields of Data Analysis, Cybersecurity and Privacy. In particular, her research focuses on utilizing AI and data analysis techniques to improve the efficiency and security of modern software systems. Her recent interest lies in recommendation system, anomaly detection and privacy-preserving through data mining and deep learning.
\end{IEEEbiographynophoto}
\vspace{-\baselineskip}
\vspace{-\baselineskip}
\begin{IEEEbiographynophoto}{Guandong Xu}
(Senior Member, IEEE) is a chair professor of artificial intelligence with the Education University of Hong Kong (EdUHK). Before joining EdUHK, he was a full professor in data science at University of Technology Sydney, Australia. His research interests include data science, recommender systems, user modeling, and social computing. He has published three monographs with Springer and CRC Press, as well as over 360 journal and conference papers. He is the editor-in-chief of the \textit{Human-Centric Intelligent Systems} and the assistant editor-in-chief of the \textit{World Wide Web Journal}. He has been serving on the editorial board or as a guest editor for several international journals. He is a fellow of the Institute of Engineering and Technology (IET) and the Australian Computer Society (ACS).
\end{IEEEbiographynophoto}

\vfill

\end{document}